\documentclass[10pt,journal,compsoc]{IEEEtran}
\usepackage{algorithm}
\usepackage{algpseudocode}
\usepackage{algorithmicx}
\usepackage{bbding}
\usepackage{subfigure}
\usepackage{graphicx}
\usepackage{amsfonts}
\usepackage{bbm}
\usepackage{booktabs}
\usepackage{balance}
\usepackage{multirow}
\usepackage{tikz}
\usepackage{amsmath}
\usepackage{color}
\usepackage{colortbl}
\usepackage{float}
\usepackage{bm}
\usepackage{enumitem}
\newtheorem{definition}{Definition}
\newcommand{\pengyu}[1]{\textcolor{black}{#1}}

\begin{document}
%
\title{Hijack Vertical Federated Learning Models As One Party}
%
%
%
%

\author{Pengyu Qiu, Xuhong Zhang\textsuperscript{\Envelope}, Shouling Ji, Changjiang Li, Yuwen Pu\textsuperscript{\Envelope}, Xing Yang, Ting Wang%
\IEEEcompsocitemizethanks{
\IEEEcompsocthanksitem P. Qiu, S. Ji, Y. Pu are with the College of Computer Science and Technology at Zhejiang University, Hangzhou, Zhejiang, 310027, China. E-mail:$\left\{qiupys,sji,yw.pu\right\}$@zju.edu.cn.\protect\\
\IEEEcompsocthanksitem X. Zhang is with the School of Software Technology at Zhejiang University, Ningbo, Zhejiang, 315048, China. E-mail: zhangxuhong@zju.edu.cn.\protect\\
\IEEEcompsocthanksitem X. Zhang and Y. Pu are also the corresponding authors of this paper.\protect\\
\IEEEcompsocthanksitem X. Yang is with the Hefei Interdisciplinary Center, National University of Defense Technology, Hefei, Anhui, 230037, China. E-mail: yangxing17@nudt.edu.cn.\protect\\
\IEEEcompsocthanksitem C. Li, T. Wang are with the College of Information Science and Technology at Pennsylvania State University, State College, PA, 16801, United States. E-mail: changjiang.li@psu.edu, inbox.ting@gmail.com.
}%
}

\IEEEtitleabstractindextext{%
\begin{abstract}
Vertical Federated Learning (VFL) is an emerging paradigm that enables collaborators to build machine learning models together in a distributed fashion. 
However, the security of the VFL model remains underexplored, particularly regarding the Byzantine Generals Problem (BGP), which is a well-known issue in distributed systems.
This paper focuses on revealing the threat of BGP in VFL systems. 
Specifically, we propose two attacks, the \textit{replay attack} and the \textit{generation attack}, to evaluate the vulnerability of VFL when there is only one malicious party. 
The goal of the adversary is to hijack the VFL model to give desired predictions. 
Moreover, considering the uneven distribution of importance among parties, we combine data poisoning with the aforementioned attacks to explore whether they can bypass the situation where the adversary has few features.
The evaluation results demonstrate the effectiveness of our attacks. 
For instance, the adversary holding only 10\% of the features can achieve an attack success rate close to 90\% on a binary classifier. 
Additionally, we evaluate potential countermeasures, and the experimental results show that their defense capability is limited and usually at the cost of performance loss of the VFL task. 
Our work highlights the need for advanced defenses to protect the prediction results of a VFL model and calls for more exploration of VFL's security issues.
\end{abstract}
}

\maketitle
\IEEEdisplaynontitleabstractindextext

%

\section{Introduction}\label{sec:intro}
The increasing importance of data as a valuable resource has brought attention to the issues of illegal data collection and data breaches, as highlighted in recent studies \cite{iwaya2018mhealth,herdiani2021analysis,fiesler2020no,afriat2020capitalism}.
This has led to a growing consensus on the need for personal data protection, and governments have implemented regulations such as the GDPR \cite{voigt2017eu} and CCPA \cite{kiselbach2012new} to define the boundaries for data collection, sharing, and transactions. 
While these measures serve to prevent malicious collection of personal data by companies, they also limit the ability of companies to leverage data for business growth.

\begin{figure}[t]
    \centering
    \setlength{\abovecaptionskip}{0.cm}
    \includegraphics[width=\columnwidth]{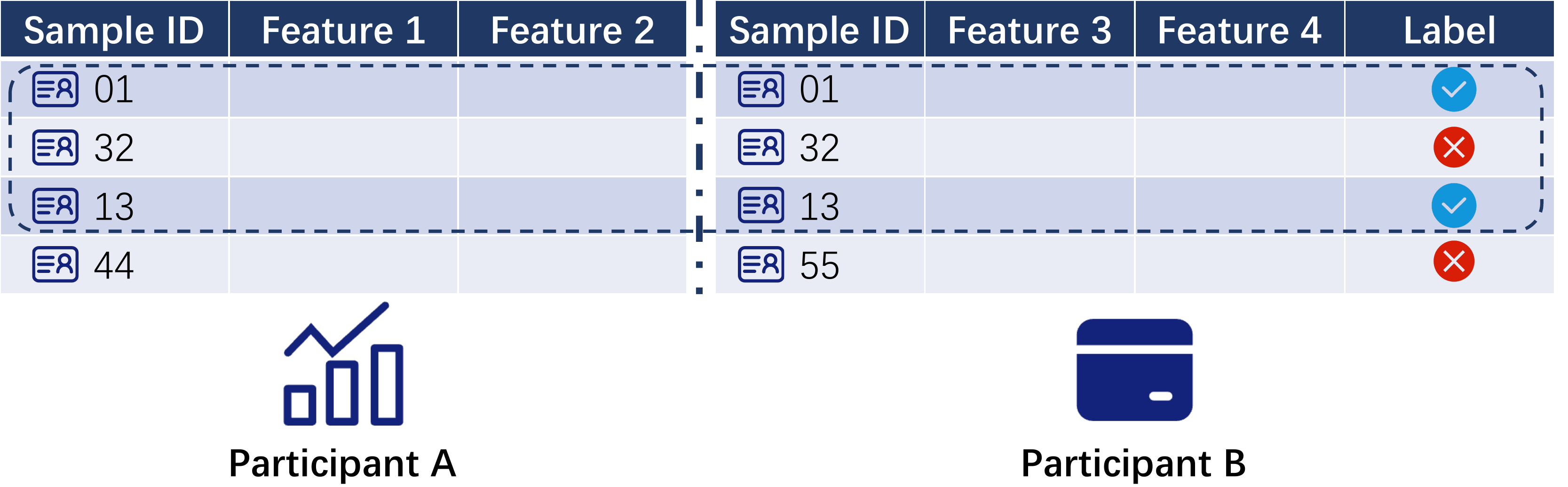}
    \caption{\pengyu{An illustration of VFL is presented. In this scenario, Participant A, a financial company, holds features 1 and 2, while Participant B, a bank, possesses features 3 and 4. The two parties collaborate to train a model for predicting loan approval.}}
    \label{fig:vfl}
\end{figure}

The challenge of balancing personal data protection with the use of data for business growth can be addressed by Vertical Federated Learning (VFL). 
VFL is specifically designed to address situations where participants hold the same samples but have different features. 
For example, in a collaboration between a bank and a financial company, as depicted in Fig.~\ref{fig:vfl}, the bank and the financial company may have a group of common users, but possess different features for those users, such as savings and loan records in the bank and stock and security information in the financial company. 
When the bank seeks to improve its loan evaluation risk control model, incorporating features from the financial company becomes a promising choice. 
VFL offers an effective solution for the bank and the financial company to train a model together.

Several frameworks, including FATE \cite{liu2021fate}, PySyft \cite{ziller2021pysyft}, TF Encrypted \cite{morten2018tfencrypted}, and CrypTen \cite{ref:crypten}, have been developed to support VFL. 
In practice, the benefits of VFL have been demonstrated in various studies such as \cite{liu2019communication,webankvflcase1,webankvflcase2}. 
However, the privacy and security analysis of VFL remains an under-explored area, as highlighted in recent research \cite{fu2022label,qiu2022relation,Luo2021feature,weng2021privacy}.

The Byzantine Generals Problem (BGP) \cite{lamport2019byzantine} is a well-known issue in the field of reliable distributed systems, highlighting the difficulty of achieving consensus in a network where some nodes may be compromised or sending false information.
In a recent study by Bagdasaryan et al. \cite{bagdasaryan2020backdoor}, a new class of backdoor attack was proposed in Horizontal Federated Learning (HFL), where participants share the same feature space but different sample spaces and collaborate to train a global model.
Their findings showed that the global model could be easily backdoored by a single party, replacing its local model with a poisoned one.

\subsection{Motivation}
\pengyu{In VFL, however, there is no complete model shared among these participants.
Furthermore, each participant holds only a portion of the user's features.
These distinctions render the straightforward application of existing attacks unsuitable. 
Nevertheless, concerns about the resilience of VFL to BGP persist, particularly regarding whether a single malicious participant might `hijack' the VFL model as can occur in HFL \cite{bagdasaryan2020backdoor}.}

\pengyu{To address this concern, our paper investigates VFLs that employ Deep Neural Networks (DNNs), which have proven highly effective in various applications \cite{krizhevsky2012imagenet,szegedy2013deep,abdel2014convolutional,tenney2019bert}.
We uncover that a single malicious participant is capable of manipulating the model to produce specific predictions by altering features. 
In order to identify susceptible features that can be exploited, we propose two attacks: the \textit{replay attack} and the \textit{generation attack}.}

\textbf{Replay Attack:} An effective attack strategy based on the intuition that robust features can be maliciously used to manipulate the model's output. 
Here, robust features refer to those strongly correlated with the labels, such as the red sports car's head and `Automobile' in CIFAR10 \cite{krizhevsky2009learning}, a widely used dataset for image classification. 
To carry out the replay attack, the adversary first selects samples that strongly align with the desired class and stores their features. 
Then, when targeting a specific sample, the adversary replaces its features on his/her end with the stored features to force the model to produce a misclassification (e.g., from `not qualified' to `pass' in a loan application in Fig.~\ref{fig:vfl}).

\textbf{Generation Attack:} The advantage of the replay attack is that it utilizes existing features, which may make it difficult to detect in cases where there are checks on data authenticity. 
However, it may not always be optimal in achieving the adversary's goal. 
The generation attack is proposed to generate stronger features than those from the replay attack, which turns the searching process into an optimization problem.

VFL has a unique property that distinguishes it from the BGP: the importance of each participant's features is highly correlated with their influence on the final prediction. 
For instance, if there are a fixed number of total features, the participant who has a larger portion of those features is assumed to have more impact on the prediction. 
This presents a challenge for an adversary who only controls a few features to perform successful attacks. 
To overcome this limitation, we propose a poisoning phase to be combined with our attacks.

\textbf{Poisoning Phase:} The poisoning phase serves to increase the VFL model's sensitivity to the adversary's features.
As described in \cite{valiant1984pac}, a model is more likely to learn the connection between the features and the target class if the features frequently appear in the samples of the class. 
Therefore, in the poisoning phase, the adversary can combine the generated features with a group of samples of the desired class to increase their frequency. 
This is supposed to result in a stronger connection between the implanted features and the class, allowing the adversary to conduct more powerful attacks.

In Section~\ref{sec:evaluation}, we present our evaluation of the proposed attacks under various scenarios. 
The experimental results demonstrate that the attacks, when combined with the poisoning phase, can achieve impressive performance. 
For instance, an adversary controlling only 10\% of the features can achieve an attack success rate of nearly 90\% on a binary classifier.

Furthermore, we also evaluate two possible countermeasures against these attacks, namely input transformation and dropout. 
However, our results indicate that the effectiveness of these defenses is limited, and they come at the cost of reduced VFL task performance.

Our contributions can be summarized as follows.
\begin{itemize}
    \item To the best of our knowledge, our work is the first to investigate the BGP in VFL with the assumption of a single malicious participant, which adds to existing knowledge on this topic.
    \item We propose two attack methods and their variants, considering the adversary's background knowledge, which are easy to initiate and have a high success rate.
    \item We evaluate our attacks on real-world datasets of various types (tabular, image, text, and multi-modal), showing their effectiveness.
    \item We analyze existing defense mechanisms and apply two selected solutions to protect against our attacks. 
    However, the experimental results indicate that more advanced defense strategies are needed. 
\end{itemize}

\section{Preliminaries}\label{sec:pre}
We begin by introducing a set of fundamental concepts and assumptions. 
TABLE~\ref{tab:notations} summarizes the important symbols and notations.

\begin{table}[ht]
\centering
\caption{Symbols and notations.}
\label{tab:notations}
\small
\resizebox{\columnwidth}{!}{%
\begin{tabular}{@{}r|p{7cm}@{}}
\multicolumn{1}{l|}{Notations} & Definition                            \\\midrule \midrule 
$P_i$, $D_i$                       & the $i$-th party in VFL, and $P_i$'s dataset \\
$\mathcal{U}_i$, $\mathcal{F}_i$   & $D_i$'s sample/user space and feature space            \\
$f_i$, $f_{top}$, $f_{s}$         &   $P_i$'s bottom model, the top model, and the surrogate model    \\
$\theta_i$, $\theta_{top}$          &    $f_i$'s parameters, and $f_{top}$'s parameters             \\
$\textbf{x}_i^{(u)}$, $y^{(u)}$, $\textbf{v}_i^{(u)}$      &  a sample $u$'s feature vector of $D_i$, $u$'s label,   $\textbf{x}_i^{(u)}$'s corresponding output from $f_i$       \\
$\mathcal{B}_{\epsilon}(\textbf{x})$ & a norm ball at $\textbf{x}$ with the radius of $\epsilon$ \\
$\textbf{x}_{\delta}$ & perturbed adversarial sample of $\textbf{x}$ \\
$\theta_\delta$ & perturbed parameter of $\theta$ \\
$d_i$, $\Tilde{d}$ & the size of features in $\mathcal{F}_i$, the output dimension of each bottom model \\
$\mathcal{P}$ & adversary's knowledge about the samples' posteriors \\
$\mathcal{L}$, $\mathcal{L}_{t}$ & adversary's knowledge about the samples' labels, adversary's knowledge about a group of samples who belongs to the target class \\
$\mathcal{K}=(\cdot,\cdot)$ & a tuple denote the adversary's background knowledge
\end{tabular}%
}
\end{table}

\subsection{Byzantine Generals Problem}
The Byzantine Generals Problem (BGP) is a well-known challenge in the design of reliable distributed systems. In their seminal work \cite{lamport2019byzantine}, Lamport et al. provided a formal description of the problem:
\begin{definition}[Byzantine Generals Problem]
A commanding general must send an order to their $n-1$ lieutenant generals such that:
\begin{itemize}
    \item IC1. All loyal lieutenants obey the same order.
    \item IC2. If the commanding general is loyal, then every loyal lieutenant obeys the order they send.
\end{itemize}
\end{definition}
Conditions IC1 and IC2 are called the interactive consistency conditions.

A fundamental conclusion of the BGP, established in \cite{lamport2019byzantine}, is that to achieve a consensus (attack or retreat) on the order in the presence of $m$ traitorous generals, the number of generals $n$ must be greater than $3m$. 
However, this conclusion does not guarantee the correctness of the consensus.

The \textit{Two Generals Paradox} is a special case of BGP that has been proven to be unsolvable \cite{lamport1985solve}.
Specifically, it presents a thought experiment in which two generals can only communicate by sending a messenger through enemy territory to agree on the time to launch an attack, but can never reach a consensus.

Our work reveals that the BGP also exists in the context of VFL. 
In addition to achieving consensus among all participants, our findings indicate that the VFL model's predictions can be easily manipulated by a single malicious party. 
Furthermore, in a two-party VFL setting, the Two Generals Paradox dilemma arises because it is impossible to determine which party's features provide the correct category information if one of them is malicious.

\subsection{Deep Neural Networks} 
Deep neural network (DNN) is one of the most popular machine learning methods in recent decades, exhibiting powerful feature extraction ability. 
Specifically, a DNN $f$ with parameters $\theta$ represents a function $f:\mathcal{X}\to\mathcal{Y}$, where $\mathcal{X}$ denotes the input space and $\mathcal{Y}$ denotes the label.

Take a classification task for example.
Given a training set $D$, of which each instance $(\textbf{x},y)\in D \subset \mathcal{X}\times\mathcal{Y}$ consists of an input $\textbf{x}$ and its class $y$, a DNN $f$ is trained to find the best parameters $\theta$ by minimizing the loss function $\ell$ (usually a cross entropy loss for classification tasks). 
Formally, the search for the best $\theta$ can be formulated as:
\pengyu{
\begin{equation}
   \min_\theta \mathbb{E}_{(\textbf{x},y)\in D}[\ell(\textbf{x},y;\theta)], 
\end{equation}
}
where $\mathbb{E}[\ell(\cdot)]$ denotes the expected loss of $f(\textbf{x};\theta)$ and $y$.

\subsection{Vertical Federated Learning}\label{sec:vfl}
Consider a classification task and a set of $N$ distributed parties $\left\{P_1,P_2,\cdots,P_N\right\}$ with datasets $\left\{D_1,D_2,\cdots,D_N\right\}$. 
For each dataset, $D_i=(\mathcal{U}_i,\mathcal{F}_i)$, where $\mathcal{U}_i$ denotes the sample space and $\mathcal{F}_i$ denotes the feature space.
Before training, VFL needs to first determine the sample space $\mathcal{U}$, which is $\mathcal{U}=\bigcap_{i=1}^{N} \mathcal{U}_i$.
Then for features from different $\mathcal{F}_i$, VFL aligns them for each sample.

After the preparation of data, $P_i$ trains its bottom model, denoted by $f_i$, 
to extract the high-level abstractions of each sample.
Let $\textbf{x}_i^{(u)}$ denote the feature vector of sample $u$ with $d_i$ features from $F_i$.
Then the function of $f_i$ is to map the feature vector to a $\Tilde{d}$-dimensional latent space, i.e., $f_i(\textbf{x}_i^{(u)};\theta_i):\mathbb{R}^{d_i}\to\mathbb{R}^{\Tilde{d}}$.
$\theta_i$ denotes the parameters of $f_i$ and is optimized by a specific objective function.
We use $\textbf{v}_i^{(u)}$ to represent the output of $f_i$.
For simplicity, unless otherwise defined, we use $\textbf{v}_i$ instead of $\textbf{v}_i^{(u)}$ in the following.

After each participant uploads $\textbf{v}_i$ to a neutral third party server (NTS), they are concatenated for further calculations.
Specifically, let $\textbf{v}_{cat}=[\textbf{v}_1,\textbf{v}_2,\cdots,\textbf{v}_N]$ and $f_{top}$ denote the top model located at NTS. 
$f_{top}$ learns a mapping from $\textbf{v}_{cat}$ to $\textbf{v}_{top}$, i.e., $f_{top}(\textbf{v}_{cat};\theta_{top}):\mathbb{R}^{N\times \Tilde{d}}\to\mathbb{R}^{c}$, where $c$ denotes the number of classes, and $\textbf{v}_{top}$ is the output of $f_{top}$, which is also the posterior of a sample.
Finally, $\textbf{v}_{top}$ is sent to the party who owns the labels to calculate the loss, e.g., cross-entropy loss.
The loss is then used to optimize each model's parameters, including $f_{top}$ and $f_i$.
Formally, the training of VFL is formulated as follows:
\pengyu{
\begin{equation}
    \min_{\left\{\theta_i\right\}_{i=1}^{N},\theta_{top}} \mathbb{E}_{u\in \mathcal{U}}[\ell(\textbf{x}_1,\textbf{x}_2,\cdots,\textbf{x}_N,y;\left\{\theta_i\right\}_{i=1}^N,\theta_{top})],
\end{equation}
}
where $\ell$ refers to the loss function.

\subsection{Threat Model}\label{sec:threat}
Assuming the adversary is a VFL participant who provides only features and aims to manipulate the VFL model into predicting a specific set of samples as the adversary's chosen class. 
The adversary may be motivated to profit by providing a service that alters the results of a user's bank loan application. 
This assumption forms the basis of the threat model.

The following are additional background knowledge of the adversary for conducting corresponding attacks.
\begin{itemize}
    \item \textbf{Access to the posteriors, denoted by $\mathcal{P}$}. 
    $\mathcal{P}$ is essential for the replay attack and generation attack.
    For the former, $\mathcal{P}$ helps identify robust features, while for the latter, it guides the optimization direction.

    The assumption of this knowledge is reasonable.
    In some consulting-like scenarios mentioned in \cite{Luo2021feature,weng2021privacy}, the adversary needs to know the detail of prediction, thus he/she can make further decisions depending on the probability distribution. 
    Moreover, according to \cite{guan2019shapley}, $\mathcal{P}$ is essential to data valuation in VFL, which determines the profit of each participant.
    Hence, there is sufficient reason for one party in VFL to access $\mathcal{P}$.
    
    \item \textbf{Auxiliary label information, denoted by $\mathcal{L}$}.
    This knowledge assumes that the adversary has access to a small portion, e.g., 1\%, of samples' ground truth labels.
    It refers to the case where $\mathcal{P}$ is unavailable, indicating the most confidential situation.

    Meeting the assumption of $\mathcal{L}$ requires additional effort. 
    As a participant, the adversary has knowledge of the samples used in training and can collect a small subset of candidate samples with unknown labels. 
    The ground truth labels can then be determined through investigation or indirect inference methods \cite{fu2022label,Liu2021BatchLI}. 
    For instance, if the adversary is a financial institution that desires knowledge of a user's credit rating, it may infer the rating by posing as a financial management inquiry and asking the user about their loan amount or total assets at the target bank.
\end{itemize}

Please note that in the poisoning phase, the adversary needs to add generated features to a group of samples from the target class.
This refers to a subset of $\mathcal{L}$.
Therefore, we further divide $\mathcal{L}$ into two cases: $\mathcal{L}_{a}$, where the adversary has knowledge of a small portion of the ground truth labels from all classes, and $\mathcal{L}_{t}$, where the adversary only knows a small portion of samples from the target class.

Since $\mathcal{P}$ and $\mathcal{L}$ are orthogonal, the adversary's background knowledge can be formally presented by a tuple $\mathcal{K}=(\cdot,\cdot)$. 
If one knowledge is unavailable, it is denoted by $\times$. 
Then, there are six different combinations of $\mathcal{K}$ (two choices for $\mathcal{P}$ and three for $\mathcal{L}$). 

When $\mathcal{K}=(\times,\times)$, the adversary is unable to conduct any attacks, and we exclude it from our consideration.
When $\mathcal{K}=(\times,\mathcal{L}_{t})$, the adversary cannot perform the replay attack or generation attack.
However, it is also an interesting question to see whether the added features alone can achieve the goal.
We leave it as a baseline in experiment, comparing to our attacks.
Then, for $\mathcal{K}=(\mathcal{P},\mathcal{L}_{a})$ and $\mathcal{K}=(\mathcal{P},\mathcal{L}_{t})$, $\mathcal{K}=(\mathcal{P},\mathcal{L}_{a})$ is excluded as $\mathcal{L}_{t}$ is sufficient for the poisoning phase.

Finally, three kinds of $\mathcal{K}$, i.e., $\mathcal{K}=(\mathcal{P},\times)$, $\mathcal{K}=(\mathcal{P},\mathcal{L}_{t})$, and $\mathcal{K}=(\times,\mathcal{L})$, remain.
Section~\ref{sec:method} presents the detailed designs and implementations of these attacks under each of the conditions.

\section{Methodology}\label{sec:method}
\subsection{Overview of Attack Pipeline}
\begin{figure}[t]
    \centering
    \setlength{\abovecaptionskip}{0.cm}
    \includegraphics[width=\columnwidth]{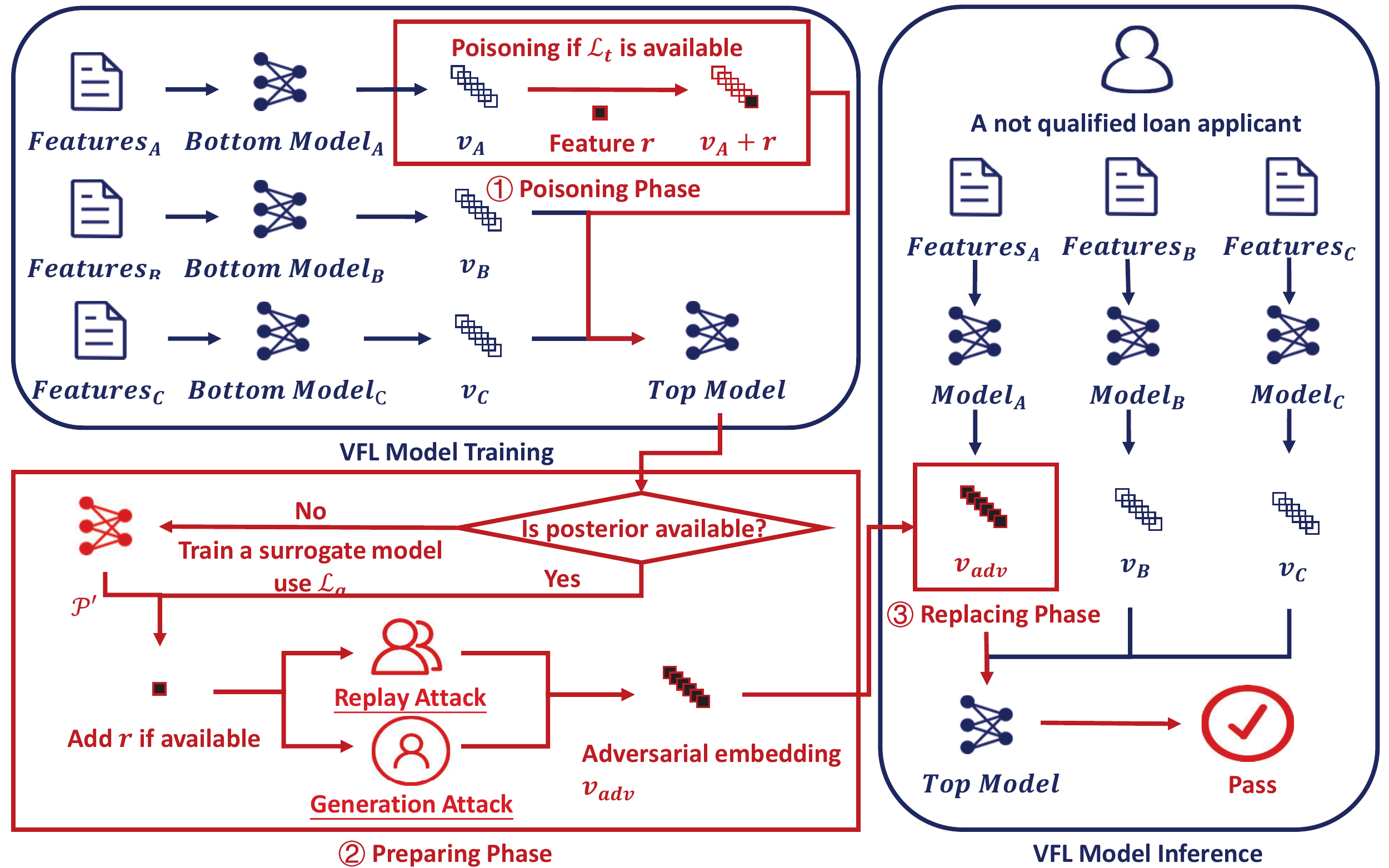}
    \caption{The overview of the attack pipeline. 1) refers to the poisoning phase if $\mathcal{K}$ contains $\mathcal{L}_{t}$. 2) refers to the preparing phase, where the adversary use $\mathcal{K}$ to conduct the replay attack and the generation attack. This part also includes the case when $\mathcal{P}$ is unavailable and the adversary uses $\mathcal{L}_{a}$ to train a surrogate model instead. 3) refers to the replacing phase, which is after preparing the embedding $\textbf{v}_{adv}$, and the adversary replaces his/her embedding with $\textbf{v}_{adv}$ for the target sample.}
    \label{fig:overview}
\end{figure}

The pipeline of our attacks is illustrated in Fig.~\ref{fig:overview} and can be divided into three phases: poisoning, preparing, and replacing. 
Since the adversary has control over his/her bottom model's output, we do not distinguish between the adversary's prepared features and the corresponding adversarial embedding in the following. 
This can also benefit our attacks because optimization in continuous space is typically easier than in discrete space.

First, the poisoning phase depends on whether $\mathcal{K}$ contains $\mathcal{L}_{t}$.
If $\mathcal{K}$ contains $\mathcal{L}_{t}$, the adversary generates random Gaussian noise $\textbf{r}$ and adds it on the embedding of each sample in $\mathcal{L}_{t}$. 
This phase aims to make the top model memorize the connection between $\textbf{r}$ and the target class, as discussed in Section~\ref{sec:intro}.

Second, after training a VFL model, the adversary should check whether the knowledge $\mathcal{P}$ is accessible.
If $\mathcal{P}$ is not accessible, the adversary should train a surrogate model using $\mathcal{L}_{a}$.
The output of the surrogate model, denoted by $\mathcal{P}'$, is then used to approximate $\mathcal{P}$.
The training of the surrogate model only uses the adversary's features, providing a view of the clustering trends of samples from the adversary's perspective.

The preparation of adversarial embeddings, $\textbf{v}{adv}$s, uses $\mathcal{P}$ (or $\mathcal{P}'$). 
Specifically, the replay attack uses $\mathcal{P}$ to identify a set of samples with high confidence in the target class. Additionally, if $\textbf{r}$ is available, it will be added to a candidate sample's embedding during the selection process. 
The embeddings of these selected samples are recorded as $\textbf{v}_{adv}$s.

In the generation attack, $\mathcal{P}$ is used to guide the generation of $\textbf{v}_{adv}$ that can change a set of samples from a certain class to the target class.
If $\textbf{r}$ is available, it will be used as the initialization for $\textbf{v}_{adv}$.
Otherwise, $\textbf{v}_{adv}$ is initialized as 0 to avoid potential interference from existing noise that may already belong to a class, which could affect the optimization process.

Finally, in the replacing phase, the adversary replaces the embedding of a target sample with $\textbf{v}_{adv}$ to achieve the desired prediction.

The details of implementations with different $\mathcal{K}$ are presented in the following sections.
However, since the replay attack is easy to implement, we do not further introduce it.

\subsection{Adversary's Knowledge: $\mathcal{K}=(\mathcal{P},\times)$}\label{sec:KP}
This knowledge refers to the scenario where the adversary acts honestly during the training phase but attempts to manipulate the model's predictions during the inference phase.
Formally, the generation attack with this knowledge can be formulated as follows:
\begin{equation}\label{eq:KP}
\mathop{\min}_{\textbf{v}_{adv}\in\mathbb{R}^{\Tilde{d}}}\ell([\textbf{v}_1,\cdots,\textbf{v}_{adv},\cdots,\textbf{v}_N],t;\left\{\theta_i\right\}_{i=1}^{N},\theta_{top}),
\end{equation}
where $\ell$ measures the entropy loss of the perturbed prediction results and the target class $t$, 
and $\left\{\theta_{i}\right\}_{i=1}^{N}$ and $\theta_{top}$ are the models' parameters. 

Solving the objective function is challenging as $\left\{\textbf{v}_i\right\}_{i\neq adv}$ and $\left\{\theta_i\right\}_{i\neq adv}$ are confidential, making it infeasible to calculate gradients for optimization.
Zeroth Order Optimization (ZOO) provides an approximation way for the gradients that only relies on the input and the output.

Given an input $\textbf{x}$ and function $f$, ZOO set up a constant $h$ and a perturbation $\bm{\delta}$ to approximate the gradients.
It can be formally described as follows:
\pengyu{
\begin{equation}
   g:=\frac{\partial f(\textbf{x})}{\partial \textbf{x}}\simeq \frac{f(\textbf{x}+h\bm{\delta})-f(\textbf{x}-h\bm{\delta})}{2h\bm{\delta}}. 
\end{equation}
}
The core idea of ZOO is to use an average slope to approximate the instantaneous slope at $\textbf{x}$.

In our case, since $\left\{\textbf{v}_i\right\}_{i\neq adv}$ and $\left\{\theta_i\right\}_{i\neq adv}$ are fixed after training, they can be regarded as a part of the unknown function.
Then, denote by $F$ the function that calculates the posteriors based on $\textbf{v}_{adv}$, i.e., $F(\textbf{v}_{adv};\left\{\textbf{v}_i\right\}_{i\neq adv},\left\{\theta_i\right\}_{i\neq adv})$, Eq~.(\ref{eq:KP}) can be reformulated as:
\pengyu{
\begin{equation}
    \mathop{\min}_{\textbf{v}_{adv}\in\mathbb{R}^{\Tilde{d}}}\left\{\ell(\textbf{v}_{adv})\right\},
\end{equation}
}
where $\ell(\textbf{v}_{adv})=\max_{i\neq t}(F(\textbf{v}_{adv})_i)-F(\textbf{v}_{adv})_t$.
The new loss function measures the gap between class $t$'s probability and the maximum probability except $t$.

To reduce the complexity of the optimization, the symmetric difference quotient \cite{lax2014calculus} is introduced to estimate the gradient at coordinate level.
Specifically, it can be formulated as follows:
\begin{equation}\label{eq:approximate}
    g_i:=\frac{\partial \ell(\textbf{v}_{adv})}{\partial \textbf{v}_{adv}[i]}\simeq \frac{\ell(\textbf{v}_{adv}+h\textbf{e}_i)-\ell(\textbf{v}_{adv}-h\textbf{e}_i)}{2h\textbf{e}_i},
\end{equation}
where $g_i$ denote the $i$-th coordinate's gradient, $\textbf{v}_{adv}[i]$ denote the $i$-th coordinate of $\textbf{v}_{adv}$, $h$ is a constant, and $\textbf{e}_i$ is a standard base vector with only the $i$-th component as 1.

However, a small perturbation in a single dimension may cause no significant change in the final output, resulting in a very small estimated gradient.
To solve the problem, we further scale up the gradient to $10^{-1}$ level if it is too small.
With the coordinate-related gradient, the following optimization process is completed by the coordinate descent method \cite{bertsekas1997nonlinear}, which subtracts the gradient at the corresponding coordinate.

Given that real-world applications, such as loan applications, often restrict users to a single application within a specified period, it is crucial to prepare the adversarial embedding vector $\textbf{v}_{adv}$ beforehand. 
However, generating a universal adversarial embedding vector that can work for all samples without the true gradients and control of all features is challenging \cite{moosavi2017universal}. 
One feasible approach is to limit $\textbf{v}_{adv}$'s ability to modify the model's predictions for samples belonging to a specific source class $s$, rather than all classes. 
By creating $\textbf{v}_{adv}$s for all source-target pairs, the adversary can achieve the goal of altering any sample's prediction. 
Furthermore, to generate the optimal $\textbf{v}_{adv}$, the small set of samples from class $s$ should have high confidence.

Algorithm~1 outlines the attack process. 
Firstly, the adversary prepares a set of samples of the source class, denoted by $\mathcal{S}$. 
Then, $\textbf{v}_{adv}$ is initialized to 0. 
This is because random initialization may result in $\textbf{v}_{adv}$ belonging to a specific class, whereas 0 has no impact on the final calculation. 
Then, the algorithm will randomly select a sample $s$ in $\mathcal{S}$ and the coordinate $i$.
By querying the top model for the outputs of $\textbf{v}_{adv}+h\textbf{e}_i$ and $\textbf{v}_{adv}-h\textbf{e}_i$, the adversary obtains an approximated gradient $g_i$.
The adversary subtracts $g_i$ from the $i$-th coordinate of $\textbf{v}_{adv}$.
The above procedure is repeated for the total query budget $Q$.
Finally, after optimization, the optimized $\textbf{v}_{adv}$ is stored for the replacing phase.

\begin{algorithm}\label{alg:KP}
    \small
    \caption{Class-specific Adversarial Embedding Generation}
    \begin{algorithmic}[1]
        \Require{Samples who have high confidence of the source class, $\mathcal{S}$; query budget $Q$.}
        \Ensure{Adversarial embedding $\textbf{v}_{adv}$.}
        \State $\textbf{v}_{adv} \gets \mathbf{0}$, $q~\gets 0$
        \While{$q<Q$}
            \State randomly select $s\in\mathcal{S}$
            \State randomly choose coordinate $i$
            \State $g_i \gets$ ZOO($\ell([\textbf{v}_1^{(s)},\cdots,\textbf{v}_{adv},\cdots,\textbf{v}_N^{(s)}],t)$)
            \State $\textbf{v}_{adv}[i] \gets \textbf{v}_{adv}[i] - g_i$
            \State $q~\gets~q+2$
        \EndWhile
    \end{algorithmic}
\end{algorithm}

\subsection{Adversary's Knowledge: $\mathcal{K}=(\mathcal{P},\mathcal{L}_{t})$}\label{sec:KPL}
When the adversary has a few features, his/her contribution to the final prediction is limited \cite{guan2019shapley}, as $\theta_{top}$ may give a small weight on $\textbf{v}_{adv}$.
This limits the adversary's capability and may cause that no matter how well $\textbf{v}_{adv}$ is optimized, it is hard to flip the prediction.
Therefore, with the help of $\mathcal{L}_{t}$, an active way to adjust the weight is perturbing $\theta_{top}$ at the training phase, i.e., the poisoning phase.

The purpose of the poisoning phase is to implant a frequently appeared feature $\textbf{r}$ in the samples with the target class $t$, hoping that the top model can `learn' the connection between $\textbf{r}$ and $t$.
The feature $\textbf{r}$ could be randomly initialized Gaussian noise.
Then, for each sample from $\mathcal{L}_{t}$, $\textbf{r}$ will be added on its output from the bottom model.

Given the poisoned VFL model, the remaining procedure is similar to the previous case.
The only difference is that $\textbf{v}_{adv}$ is initialized as $\textbf{r}$ instead of 0.
The reason is that searching in the neighborhood of $\textbf{r}$ can improve the generation efficiency, as the top model is supposed to be sensitive to $\textbf{r}$.
The objective function is updated to:
\begin{equation}\label{eq:kpl}
\mathop{\min}_{\textbf{v}_{adv}\in\mathbb{R}^{\Tilde{d}}}\ell([\textbf{v}_1,\cdots,\textbf{v}_{adv},\cdots,\textbf{v}_N],t;\left\{{\theta_\delta}_i\right\}_{i=1}^{N},{\theta_\delta}_{top}),
\end{equation}
where $\left\{{\theta_\delta}_i\right\}_{i=1}^{N}$ and ${\theta_\delta}_{top}$ denote the perturbed parameters. 
Algorithm~2 summarizes the process of the attack.

\begin{algorithm}\label{alg:KPL}
    \small
    \caption{Adversarial Embedding Generation with Poisoning Attack}
    \begin{algorithmic}[1]
        \Require{Auxiliary set of samples with target label $\mathcal{L}_{t}$; samples who have high confidence of the source class $\mathcal{S}$; query budget $Q$.}
        \Ensure{Adversarial embedding $\textbf{v}_{adv}$.}
        \State \textbf{Poisoning Phase:}
        \State Generate Gaussian noise $\textbf{r}$.
        \State Add $\textbf{r}$ on the samples' embeddings from $\mathcal{L}_{t}$, and upload them in training.
        \State \textbf{Preparing Phase:}
        \State $\textbf{v}_{adv} \gets \textbf{r}$, $q~\gets 0$
        \State The following procedure is the same as Algorithm~1.
    \end{algorithmic}
\end{algorithm}

\subsection{Adversary's Knowledge: $\mathcal{K}=(\times,\mathcal{L}_{a})$}\label{sec:KL}
In this scenario, the adversary's knowledge is limited to the auxiliary label information $\mathcal{L}_{a}$ and does not have access to feedback from the VFL model. 
His/her role is a participant who is paid to provide features.

Although the adversary lacks access to $\mathcal{P}$, he/she can still perturb the top model during training by following the poisoning method outlined in the previous section. 
Then, to overcome the challenge of lacking $\mathcal{P}$, $\mathcal{L}_{a}$ is used to train a surrogate model $f_s$.

Specifically, $f_s$ takes the output of the adversary's bottom model as input (with $\textbf{r}$ added if the sample belongs to the target class) and aims to approximate the classification boundary of $f_{top}$ using only the adversary's features.

To train a high-quality surrogate model with a small number of samples, mix-up is employed in $f_s$'s training. 
Mix-up, proposed by Zhang et al. in \cite{zhang2018mixup}, implements data augmentation through combinations of samples from different classes. 
For example, given two samples $u$ and $v$, mix-up creates a new sample $\textbf{x}_{new} = \lambda \textbf{x}^{(u)} + (1-\lambda) \textbf{x}^{(v)}$, with a new label $y_{new}=\lambda y^{(u)} + (1-\lambda) y^{(v)}$.
Compared to conventional techniques such as rotation, clipping, and scaling, mix-up has demonstrated significant improvements on original tasks \cite{zhang2018mixup}.

However, $f_s$ still faces the challenge of transferability of the generated $\textbf{v}_{adv}$ between $f_s$ and $f_{top}$.
This is because much information is not accessible, i.e., $\left\{\textbf{v}_i\right\}_{i\neq adv}$ and $\left\{\theta_i\right\}_{i\neq adv}$, making it difficult to evaluate whether the generated $\textbf{v}_{adv}$ can overcome the influence of other participants' features.
To address this challenge, we focus on strengthen $\textbf{v}_{adv}$'s signal on the target class.
The attack can be formulated as follows:
\begin{equation}\label{eq:kl}
     \mathop{\min}_{\textbf{v}_{adv}\in\mathbb{R}^{\Tilde{d}}}\ell(f_s(\textbf{v}_{adv}),t).
\end{equation}

Algorithm 3 outlines the attack process. 
First, the adversary conducts the poisoning phase using the subset $\mathcal{L}_t$ of $\mathcal{L}_a$. 
Next, the adversary trains a surrogate model $f_s$ using $\mathcal{L}_a$ and their features, with mix-up used for data augmentation.
After $f_s$'s training, $\textbf{v}_{adv}$ is initialized with a sample $u$ that has high confidence with the target class. 
The optimization process then strengthens the signal of $\textbf{v}_{adv}$ on the target class until the loss is within the given bound. 

\begin{algorithm}\label{alg:KL}
    \small
    \caption{Adversarial Embedding Generation with Auxiliary Label Information}
    \begin{algorithmic}[1]
        \Require{Auxiliary label information with all classes $\mathcal{L}_{a}$, error bound $\epsilon$.}
        \Ensure{Adversarial embedding $\textbf{v}_{adv}$.}
        \State \textbf{Poisoning Phase:}
        \State Generate Gaussian noise $\textbf{r}$.
        \State Add $\textbf{r}$ on embeddings of samples with target class from $\mathcal{L}_{a}$, and upload them at training time.
        \State \textbf{Preparing Phase:}
        \State Train a surrogate model $f_s$.
        \State Collecting a sample $u$ with high confidence of target class through $f_s$.
        \State $\textbf{v}_{adv} \gets \textbf{v}_{adv}^{(u)}+\textbf{r}$.
        \While{$\ell(f_s(\textbf{v}_{adv}),t)>\epsilon$}
            \State $\textbf{v}_{adv} \gets \textbf{v}_{adv} - \frac{\partial\ell(f_s(\textbf{v}_{adv}),t)}{\partial\textbf{v}_{adv}}$
        \EndWhile
    \end{algorithmic}
\end{algorithm}

\section{Experimental Setup}\label{sec:setup}
Based on \cite{Luo2021feature,fu2022label}, we focus on evaluating two-party VFL as the primary case, while analyzing the impact of the number of parties in Section~\ref{sec:sensitivity}. 
Real-world cases demonstrated by FedAI \cite{webankvflcase1,webankvflcase2} and previous works of VFL \cite{yang2019federated,Luo2021feature,zhou2021vertically} suggest that two-party VFL is more commonly used than multi-party VFL. 
We speculate that this is due to a decrease in overlapped users among participating companies and an increase in communication costs as the number of parties increases.

\subsection{Datasets}
We evaluate the attack performance on 8 public datasets, including tabular, image, text, and multi-modal data types:
1) Company Bankruptcy Prediction Dataset (denoted by BANK) \cite{LIANG2016561}, which was collected from the Taiwan Economic Journal for the years 1999 to 2009, consisting of 6,819 instances with 95 attributes and two classes;
2) CRITEO \cite{Guo2017criteo}, which is used for Click-Through-Rate (CTR) prediction, consisting of 100,000 instances;
3) MNIST \cite{lecun1998gradient}, which is widely used as the benchmark in machine learning, consisting of 60,000 images with 10 classes; 
4) CIFAR10 \cite{krizhevsky2009learning}, which is also a well-known image dataset, consisting of 60,000 images with 10 classes;
5) CIFAR100 \cite{krizhevsky2009learning}, which shares the same images as CIFAR10 but has 100 classes;
6) Facial Expression Recognition (FER) \cite{Barsoum2016fer}, which consists of a training set of 28,709 examples and a test set of 7178 examples;
7) IMDB \cite{maas2011word}, which contains 50,000 movie reviews with two classes; and 
8) Android Permission Dataset (denoted by AP) \cite{arvind2018android}, 
which contains 20,000 android APPs' permission statistics and descriptions with two classes.

For BANK, CRITEO, and AP, we further use the oversampling method to balance the number of samples in each category.
Then the train and test data is split with a ratio of 7:3.
As for MNIST, CIFAR10, FER, and IMDB, they are split by default.
Specifically, CIFAR10 uses 50,000 images in the training and 10,000 images for testing. 
IMDB, however, uses 25,000 reviews in training and 25,000 reviews for testing.

\textbf{Dataset Configuration:}
To simulate a two-party VFL scenario, we adopt the approach described in \cite{Luo2021feature,fu2022label} and divide the features into two parts. 
For tabular data, splitting it into two parts is straightforward. 
When dealing with images, we split them vertically along the center, meaning that if the feature ratio is 50\%, we divide the image into two halves. 
As for text data, we split it based on the proportion of words.

In our experiments, we also vary the feature ratio of the adversary from 10\% to 90\% to approximate his/her feature importance to the VFL model. 
Here, the feature ratio refers to the proportion of features owned by the adversary. 
According to \cite{guan2019shapley}, the contribution of each party to the prediction results in VFL is proportional to the number of features they own. 
Therefore, using the feature ratio to represent the importance of the adversary is reasonable.

\subsection{Models and Default Parameters}
In our experiments, we train the VFL model in a local environment and designate one bottom model as the adversary's model. 
TABLE~\ref{tab:default} provides an overview of the models and other default parameters.

\textbf{Models:} The top model comprises dense layers, including an input layer, an output layer, and one hidden layer. The bottom model is chosen based on the type of data. 
For tabular data, we use MLP \cite{Tang2016ExtremeLM} with three dense layers to compute embeddings. 
For image data, we use ResNet \cite{he2015deep}, while for text data, we use BERT \cite{devlin2019bert} as the bottom model. 
The surrogate model is also made up of dense layers.

\textbf{Default Parameters:} 
We train the VFL model for 10 epochs, as pre-trained models from PyTorch \footnote{https://pytorch.org} and Hugging Face \footnote{https://huggingface.co} are employed as bottom models. 
Each batch for tabular and image data consists of 256 samples, while for text data, it is limited to 8 due to memory constraints.

During the poisoning phase, only 1\% of samples from the target class are used, which is comparable to the number used in conventional poisoning attacks \cite{ali2018poison}. 
Moreover, we limit the query budget of the generation attack to 300 iterations. 
These stringent settings take into account the limitations of the practical implementation in the real world, such as the cost of selecting users' information, and help to expose the serious security issues in VFL.

\begin{table}[t]
\centering
\caption{Models and default parameter settings.}
\label{tab:default}
\small
\resizebox{\columnwidth}{!}{%
\begin{tabular}{p{2.4cm}|p{3.6cm}|p{3cm}}
Term                            & Parameter                 & Setting                     \\ \midrule \midrule
\multirow{2}{*}{MLP (Bottom)}    & size of hidden layer      & width=64          \\
                                & size of output dimension  & $\Tilde{d}=16$       \\ \hline
\multirow{2}{*}{MLP (Top)}       & size of hidden layer      & width=64            \\
                                & size of input layer        & in\_features=$16N$  \\ \hline
\multirow{2}{*}{MLP (Surrogate)} & size of hidden layer      & width=64            \\ 
                                & size of input layer        & in\_features=16      \\ \hline
\multirow{2}{*}{ResNet}         & number of block layers     & depth=18            \\
                                & size of output dimension         & $\Tilde{d}=16$         \\ \hline
BERT                            & type                             & bert-base-uncased \\ \midrule \midrule
\multirow{2}{*}{Training}           & epochs                 & 10 \\
                                    & batch\_size             & 256 (images and tabular data) or 8 (texts) \\ \hline
\multirow{2}{*}{Adam}           & learning rate                    & $lr=1e-3$         \\
                                & weight decay                     & $wd=5e-4$    \\ \hline
\multirow{2}{*}{Poisoning Phase}    & time to poison samples & after epoch=5                   \\
                                    & poison ratio           & 1\%, e.g., 50 for CIFAR10 \\ \hline
\multirow{2}{*}{ZOO}                & height                 & h=0.01                 \\
                                    & step                   & step=0.01                  \\ \hline
\multirow{2}{*}{Mix-up}             & training epoch         & epoch=300                \\
                                    & random seed            & $\alpha=1$                 
\end{tabular}%
}
\end{table}

\subsection{Metrics}
The metrics used for evaluating our attacks includes the Attack Success Rate (ASR), Running Time (RT), and Memory Usage (MU).

\textbf{ASR:} The main metric used to evaluate the attack performance, which is defined as follows:
\pengyu{
\begin{equation}
    \frac{\sum\limits_{u \in \mathcal{D}_{test}}I(\mathop{\arg\max}(f(\textbf{v}_1^{(u)},\cdots,\textbf{v}_N^{(u)};\theta_{top}))=t)}{\left\|\mathcal{D}_{test}\right\|},
\end{equation}
}
where $I(\cdot)$ is an indicator function and $\mathcal{D}_{test}$ denotes the testing dataset.
For both the replay attack and the generation attack, ASR is tested with 10 $v_{adv}$s and the averaged results is reported.

\textbf{RT:} The evaluation of RT is straight forward.
It records the timestamps from the start of the program to the end, and then records the time cost by the attack.

\textbf{MU:} The MU is recorded during the process and is implemented by \textit{memory\_profile} and \textit{psutil}, which are prevalent packages used to record process and system utilities.

\subsection{Baselines}\label{sec:baseline}
We introduce two baseline methods for comparison. 
The first is the \textit{poison attack}, which corresponds to the scenario where $\mathcal{K}=(\times,\mathcal{L}_t)$. This baseline evaluates whether the frequently implanted feature $\textbf{r}$ can serve as a `trigger' similar to that in backdoor attacks \cite{ali2018poison}. 
The poison attack is easy to implement, with the adversary adding $\textbf{r}$ to their embeddings.

The second baseline is a modification of the backdoor attack in HFL \cite{bagdasaryan2020backdoor}, called the \textit{scale attack}. 
In \cite{bagdasaryan2020backdoor}, a poisoned model is trained, and its perturbed parameters are scaled before the aggregation process.
Following their approach, the scale attack selects a sample from the target class randomly and scales its embedding, such as by a factor of 10, before uploading it to the top model.

\subsection{Environment}
We implement the attacks in Python and conduct all experiments on a workstation equipped with AMD Ryzen 9 3950X and an NVIDIA GTX 3090 GPU card. 
We use PyTorch to implement the models used in the experiments, and pandas and sklearn for data preprocess.
\section{Evaluation}\label{sec:evaluation}
In this section, we present three different scenarios based on $\mathcal{K}$. 
The purposes of each scenario are organized as follows:
\begin{itemize}
    \item In Section~\ref{sec:ap}, we evaluate our attacks in the basic setting when $\mathcal{K}=(\mathcal{P},\times)$.
    \item In Section~\ref{sec:apt}, we conduct experiments to evaluate the effect of combining the poisoning phase when $\mathcal{K}=(\mathcal{P},\mathcal{L}_{t})$.
    \item In Section~\ref{sec:al}, we evaluate whether the adversary alone can conduct the attacks when $\mathcal{K}=(\times,\mathcal{L}_{a})$.
\end{itemize}

Given that we limit the scope of $\textbf{v}_{adv}$ to modifying a set of samples from a specific class, our evaluations and comparisons focus on a single fixed source class. 
Specifically, we use `0' as the source class, which corresponds to `bankrupt' in BANK, `airplane' in CIFAR10, `negative' in IMDB, and so on. 
For binary classification tasks such as BANK, CRITEO, and IMDB, we set the target class to 1'. 
Likewise, for image datasets like CIFAR10, we also set the target class as 1', which corresponds to the `automobile' class, to ensure consistency across experiments. 
We analyze the impact of different combinations of source and target classes in Section~\ref{sec:pairs}.

\subsection{Evaluation with $\mathcal{K}=(\mathcal{P},\times)$}\label{sec:ap}
In this section, we evaluate the effectiveness of our attacks using $\mathcal{K}=(\mathcal{P},\times)$, and address the following questions:
\begin{itemize}
    \item How effective are our attacks against a normal VFL model?
    \item Can the adversarial embedding successfully change the prediction of a sample to the target class?
    \item Do our attacks work well across all types of datasets?
\end{itemize}

\begin{figure}[t]
    \centering
    \setlength{\abovecaptionskip}{0.cm}
    \includegraphics[width=\columnwidth]{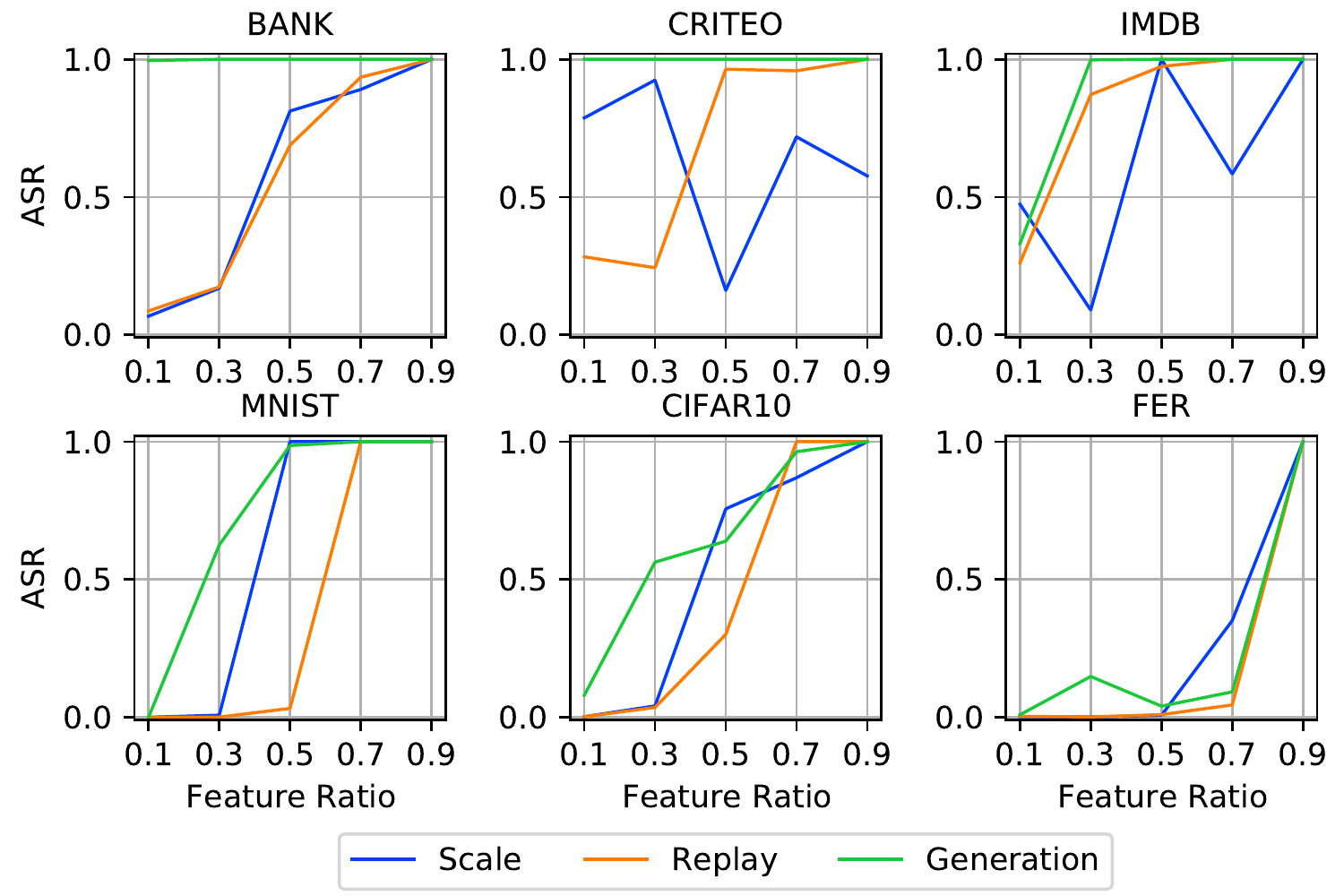}
    \caption{Attack performance with $\mathcal{K}=(\mathcal{P},\times)$. The x-axis represents the feature ratio of the adversary. The y-axis represents the ASR.}
    \label{fig:ap}
\end{figure}

\textbf{Attack Performance:} Fig.~\ref{fig:ap} illustrates the results of our attacks with $\mathcal{K}=(\mathcal{P},\times)$.
For BANK, CRITEO, and IMDB, we observe that the generation attack can achieve remarkable performance with an ASR of nearly 99\%, given a feature ratio higher than 30\%.
For MNIST and CIFAR10, the ASR becomes considerable when the adversary has over 50\% features.
However, for FER, the ASR only reaches 99\% when the feature ratio is set to 90\%.

We speculate that the difference in the success rate between the tabular/text dataset and image datasets is due to the completeness of features.
For example, in BANK, the feature split distributes different attributes to each party, but it does not split the attribute values themselves.
On the other hand, in MNIST, the feature split means that the image of a number will be cut into two parts, which can make our attacks more challenging to succeed as the adversary does not have valid and complete features to work with.

Regarding the difference between FER and MNIST/CIFAR10, we speculate that the performance of the main task may contribute to the observed phenomenon. 
In particular, the classification accuracy of FER on the test set is only around 60\%, whereas MNIST and CIFAR10 achieve over 90\% accuracy. 
This means that even true data may be misclassified by the model in the case of FER, which could reduce the success rate of our attacks.

The baseline, scale attack, and replay attack are effective when the feature ratio exceeds 50\%. 
In binary tasks, the replay attack demonstrates greater stability compared to the scale attack, particularly on the CRITEO dataset. 
However, for multi-label tasks, the scale attack outperforms the replay attack.

We hypothesize that the classification boundary for multi-label tasks is more complex than that of binary tasks. 
As a result, in binary tasks, merely scaling the values will not immediately achieve the desired outcome; it is necessary to identify features that cross the boundary.
In contrast, for tasks with complex boundaries, features significant for one class might also be important for another class. 
This implies that replacing features could lead to incorrect predictions instead of the targeted class. 

In conclusion, the above results demonstrate that the success of our attacks is independent of the data type, as expected.
Moreover, for tabular and text data, an attribute or a word may be decisive in determining the class, whereas for images, a salient complete object, such as cat ears, is required.

\subsection{Evaluation with $\mathcal{K}=(\mathcal{P},\mathcal{L}_t)$}\label{sec:apt}
In this section, our goal is to verify speculations as follows.
\begin{itemize}
    \item The poison attack alone cannot work with a limited number of poisoning samples;
    \item The perturbed model leads to a better performance of two attacks.
\end{itemize}

\textbf{Configuration of $\mathcal{L}_t$:}
We randomly sample auxiliary label information from the training dataset, with a sampling rate of 1\%. 
For example, in CIFAR10, only 50 images of the target class are poisoned. 
Our setting is more strict for the adversary compared to conventional poisoning attack settings \cite{ali2018poison}.

\begin{figure}[t]
    \centering
    \setlength{\abovecaptionskip}{0.cm}
    \includegraphics[width=\columnwidth]{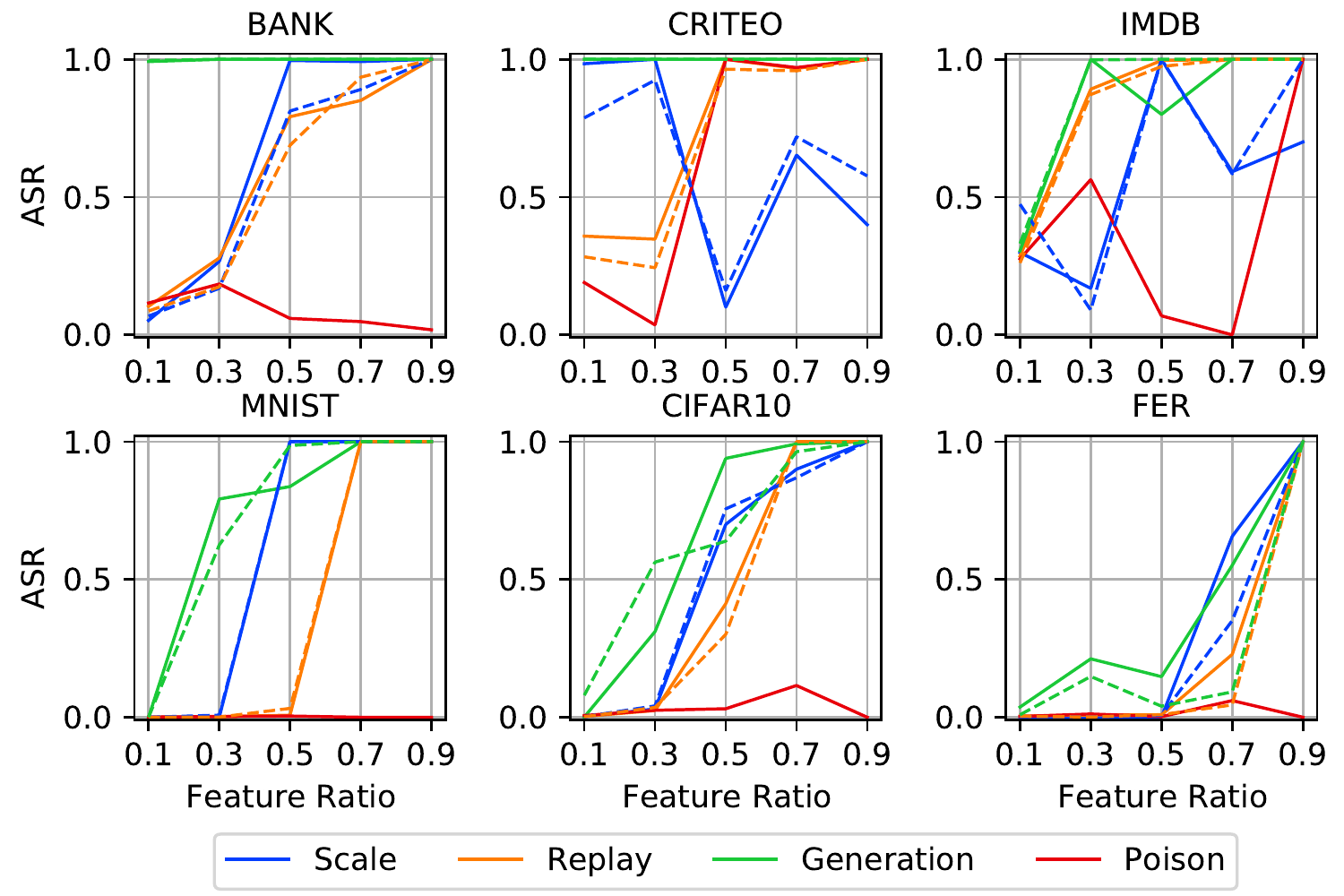}
    \caption{Attack performance with $\mathcal{K}=(\mathcal{P},\mathcal{L}_t)$. The x-axis represents the feature ratio of the adversary. The y-axis represents the ASR. \pengyu{The solid line refers to the attack performance with $\mathcal{K}=(\mathcal{P},\mathcal{L}_t)$, while the dashed line represents the corresponding attack performance with $\mathcal{K}=(\mathcal{P},\times)$.}}
    \label{fig:apt}
\end{figure}

\textbf{Attack Performance:}
The results of our attacks are presented in Fig.~\ref{fig:apt}. 
Our evaluation reveals that the poison attack alone is not effective, as its purpose is to attract the top model's attention rather than inducing a wrong prediction. 
However, for CRITEO and IMDB, even the poison attack can achieve a high ASR when the feature ratio is over 50\%, indicating that the top models trained on these datasets may be more sensitive to the feature values.

Overall, our attacks benefit from the poisoning phase. 
For example, for CIFAR10, when the feature ratio is 50\%, the generation attack's performance improves from nearly 60\% to over 90\% with the help of the poison attack. 
Similarly, the replay attack's performance improves from 30\% to 47\% when the ratio is 50\%. 
The improvements are also observed on other datasets.

These results demonstrate the effectiveness of our proposed attacks when the adversary provides valid features. 
In such cases, the attacks can achieve remarkable performance, highlighting the severe threat posed by a malicious party in VFL.

\subsection{Evaluation with $\mathcal{K}=(\times,\mathcal{L}_a)$}\label{sec:al}
In this section, we assess the performance of our attack when $\mathcal{K}$ is limited to $\mathcal{L}_a$. 
Our objective is to discover or generate embeddings that exhibit strong signals related to the target class under this stringent condition.

\textbf{Configuration of $\mathcal{L}_a$:}
We also sample $\mathcal{L}_a$ from the training dataset, with a sampling ratio of 1\%. 
For CIFAR10, this results in a total of only 500 images, with 50 images per class. 
This setup poses a significant challenge to train our mix-up based surrogate model.

\begin{figure}[t]
    \centering
    \setlength{\abovecaptionskip}{0.cm}
    \includegraphics[width=\columnwidth]{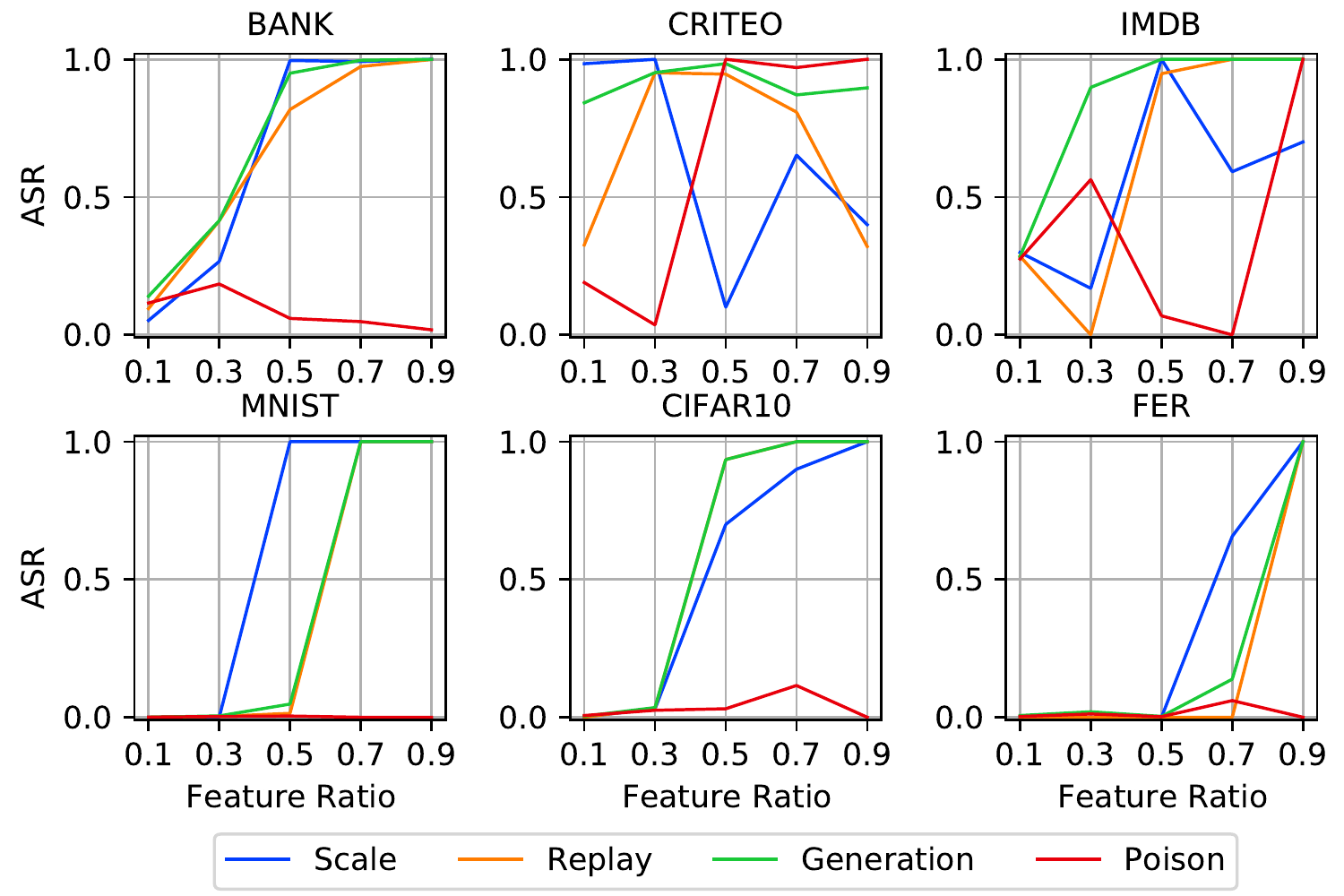}
    \caption{Attack performance with $\mathcal{K}=(\times,\mathcal{L}_a)$. The x-axis represents the feature ratio of the adversary. The y-axis represents the ASR.}
    \label{fig:al}
\end{figure}

\textbf{Attack Performance:}
The results of our attacks are presented in Fig.~\ref{fig:al}. 
In comparison to the previous scenarios, the performance of our generation attacks decreases. 
This outcome is expected, as we lack global information to optimize the adversarial embedding and instead focus on maximizing the confidence given by the surrogate model. 
Therefore, the loss of information introduced by the surrogate model and the potential overfitting of the embedding on the surrogate model contribute to the performance drop.

Nevertheless, our attacks remain potent when the feature ratio is over 50\%, and even 30\% for binary tasks.
We speculate that this is because the adversary places significant importance on the final prediction with high feature ratios, and the clustering tendency of the adversary's embedding is captured by the surrogate model \cite{fu2022label,qiu2022relation}. 

In conclusion, when $\mathcal{K}$ is limited to $\mathcal{L}_a$, our attacks are affected, but they can still perform well when the adversary has enough features that contribute to the final prediction.

\section{Sensitivity Analysis}\label{sec:sensitivity}
In this section, we further evaluate the impact of the default settings used in the previous experiments. 
We aim to answer the following research questions:
\begin{itemize}
    \item How does the number of parties affect our attacks?
    \item Will more classes invalidate our attacks?
    \item Does the choice of source and target classes affect our attacks?
    \item What is the impact of poisoning phase on our attacks?
    \item How does query budget affect the generation attack?
    \item What is the impact of multi-modal data on our attacks?
    \item What is the complexity and running time of our attacks?
    \item How much memory does our attacks take?
\end{itemize}

\subsection{Number of Parties}\label{sec:mp}
In this section, we assess the impact of the number of parties on our attacks, with a fixed adversary feature ratio of 10\%. 
Each newly added participant is assigned 10\% of the total features. 
For instance, when there are three participants, both the adversary and the newly joined participants have a feature ratio of 10\%, and the participant with labels owns the remaining 80\%. 
When there are ten participants, each one has 10\% of the features. 
The typical datasets BANK and CIFAR10 are evaluated in the following.

\begin{figure}[t]
    \centering
    \setlength{\abovecaptionskip}{0.cm}
    \includegraphics[width=\columnwidth]{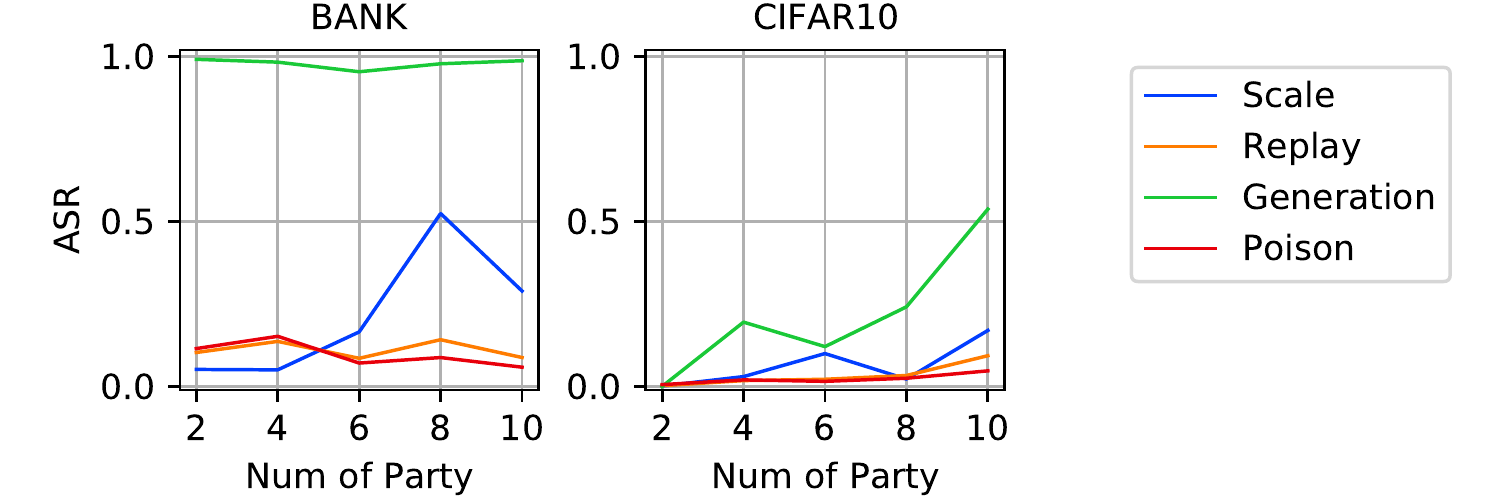}
    \caption{Attack performance with different number of parties. The x-axis represents the number of parties. The y-axis represents the ASR.}
    \label{fig:mp}
\end{figure}

The results are summarized in Fig.~\ref{fig:mp}. 
We observe that increasing the number of parties does not decrease the performance of our attacks. 
On the contrary, the attacks' performance even increases when there are more parties. 
We speculate that this is because when the features are distributed among more participants, the adversary gains more influence over the final predictions. 
For example, when there are only two parties, the adversary is the weaker one with only 10\% of the features, and the other party dominates the predictions. 
However, when there are ten parties, each participant holds 10\% of the features, giving the adversary more opportunities to manipulate the model with their prepared $\textbf{v}_{adv}$s.

In conclusion, when there are multiple parties, and each one contributes evenly to the final predictions, the adversary is more likely to manipulate the final results as a single party.

\subsection{Number of Classes}\label{sec:nc}
In this section, we evaluate the performance of our attacks with a larger number of classes. 
Our objective is to investigate the impact of the number of classes on our attacks, as complex classification boundaries may make generating $\textbf{v}_{adv}$ challenging. 
We conduct the experiments using CIFAR10 and CIFAR100 datasets. 
The attacks are performed with $\mathcal{K}=(\mathcal{P},\mathcal{L}_t)$, where the adversary's ability is maximum.

\begin{figure}[t]
    \centering
    \setlength{\abovecaptionskip}{0cm}
    \includegraphics[width=\columnwidth]{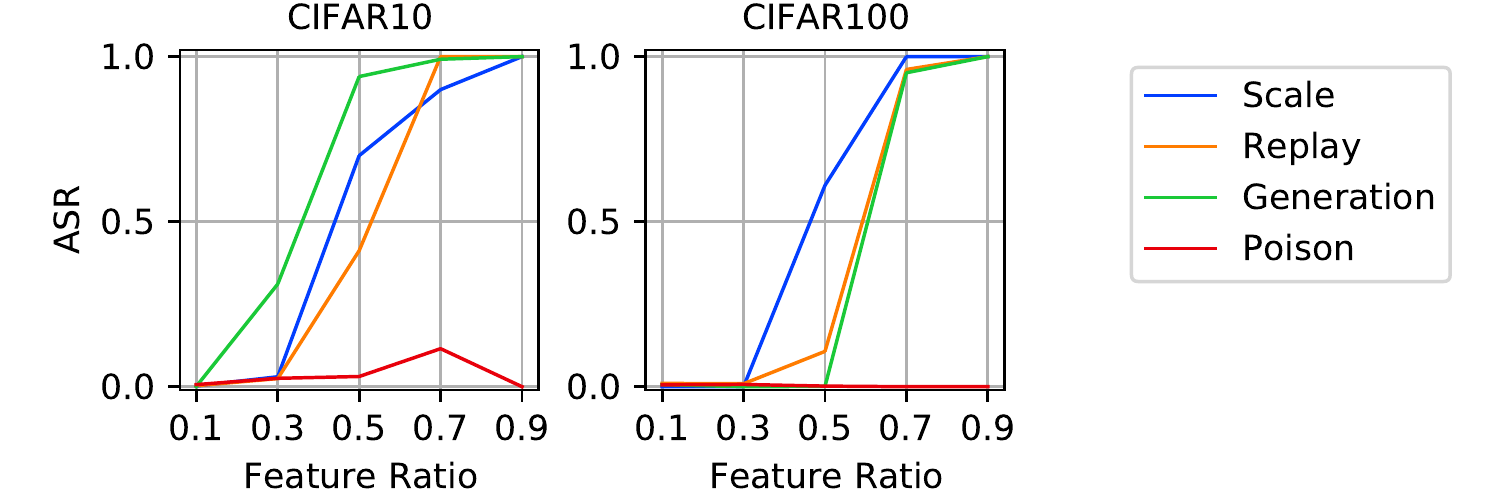}
    \caption{Attack performance when the number of classes changes. The x-axis denotes the feature ratio of the adversary. The y-axis denotes the ASR.}
    \label{fig:nc}
\end{figure}

The results for both datasets with different feature ratios are shown in Fig.~\ref{fig:nc}. 
The performance for CIFAR10 has been introduced in Section~\ref{sec:apt}, here we do not repeat it.
As for CIFAR100, all attacks' performances are affected when the feature ratio is less than 70\%. 
This outcome is reasonable as complex classification requires more diverse embeddings from both parties, i.e., more clustering centers.
Consequently, with the same feature ratio, it is more challenging for the adversary to flip the prediction without sufficient influence.

In conclusion, the experiments indicate that our attacks may experience a loss of performance as the number of classes increases. 
Nonetheless, our attacks still function effectively when dealing with 100 classes.

\subsection{Impact of Different Pairs}\label{sec:pairs}
Our previous evaluation is conducted with a fixed source-target pair by default. 
Therefore, it is important to evaluate our attacks under different source-target pairs. 
We use CIFAR10 to perform the following experiments. 
Since our attacks do not perform well with a feature ratio of 10\%, we set the feature ratio as 30\% for comparison. 
In this section, we focus on the generation attack.

Table~\ref{tab:pairs} summarizes the results, which show that the performance of our attacks varies with different source and target pairs. 
In general, our attacks perform better when the source and target classes are both animals or vehicles. 
For example, to change the prediction from `Horse' to `Deer', our generation attack can achieve an ASR of 87.23\%. 
This is reasonable since these classes are similar in our human concept, and the model can capture the relationship between them as expected.

Moreover, we find that the relationship also exists in a fixed source and target pair. 
For example, if we generate a $\textbf{v}_{adv}$ to change the prediction from `Dog' to `Cat', the model may predict `Deer' for a set of samples. 
This indicates that in VFL, compared to a targeted attack, an untargeted attack is easier to perform as it does not require any extra design.

To summarize, the difficulty of the attacks varies with different pairs. 
If the two concepts are similar and can be distinguished by the model, then it is easier for the adversary to launch a successful attack.

\begin{table}[t]
\centering
\caption{Performance of different source and target pairs. The row denotes the source classes, and the column denotes the target classes.}
\label{tab:pairs}
\resizebox{\columnwidth}{!}{%
\begin{tabular}{@{}l|cccccccccc@{}}
\toprule
 &
  \multicolumn{1}{l}{Airplane} &
  \multicolumn{1}{l}{Automobile} &
  \multicolumn{1}{l}{Bird} &
  \multicolumn{1}{l}{Cat} &
  \multicolumn{1}{l}{Deer} &
  \multicolumn{1}{l}{Dog} &
  \multicolumn{1}{l}{Frog} &
  \multicolumn{1}{l}{Horse} &
  \multicolumn{1}{l}{Ship} &
  \multicolumn{1}{l}{Truck} \\ \midrule
Airplane   & -       & \textbf{56.74\%} & 34.51\% & 18.18\%          & 7.08\%           & 8.19\%           & 5.38\%  & 1.02\%  & 41.84\%          & 5.61\%  \\
Automobile & 12.03\% & -                & 12.57\% & 2.24\%           & 0.42\%           & 2.01\%           & 0.38\%  & 8.34\%  & \textbf{40.85\%} & 26.73\% \\
Bird       & 40.23\% & 10.33\%          & -       & 19.67\%          & \textbf{59.01\%} & 12.24\%          & 13.66\% & 14.66\% & 24.73\%          & 0.47\%  \\
Cat        & 1.43\%  & 0.64\%           & 32.79\% & -                & \textbf{68.86\%} & 49.89\%          & 11.08\% & 4.11\%  & 21.10\%          & 5.05\%  \\
Deer       & 6.25\%  & 0.10\%           & 56.87\% & 8.68\%           & -                & \textbf{57.13\%} & 24.91\% & 50.81\% & 18.99\%          & 43.35\% \\
Dog        & 5.69\%  & 0.05\%           & 3.41\%  & \textbf{70.51\%} & 56.13\%          & -                & 3.12\%  & 19.46\% & 0.86\%           & 1.02\%  \\
Frog       & 7.38\%  & 0.91\%           & 44.74\% & \textbf{60.39\%} & 47.14\%          & 13.03\%          & -       & 0.08\%  & 22.72\%          & 13.96\% \\
Horse      & 0.51\%  & 0.01\%           & 5.64\%  & 16.03\%          & \textbf{87.23\%} & 20.16\%          & 1.42\%  & -       & 0.82\%           & 7.80\%  \\
Ship       & 2.49\%  & \textbf{79.00\%} & 2.25\%  & 15.22\%          & 24.10\%          & 15.75\%          & 10.17\% & 0.01\%  & -                & 24.15\% \\
Truck      & 55.37\% & \textbf{68.24\%} & 13.43\% & 35.31\%          & 6.30\%           & 27.60\%          & 31.52\% & 2.99\%  & 18.57\%          & -       \\ \bottomrule
\end{tabular}%
}
\end{table}

\subsection{Impact of Poisoning Phase}\label{sec:poison}
\begin{figure}[t]
    \centering
    \setlength{\abovecaptionskip}{0.cm}
    \includegraphics[width=\columnwidth]{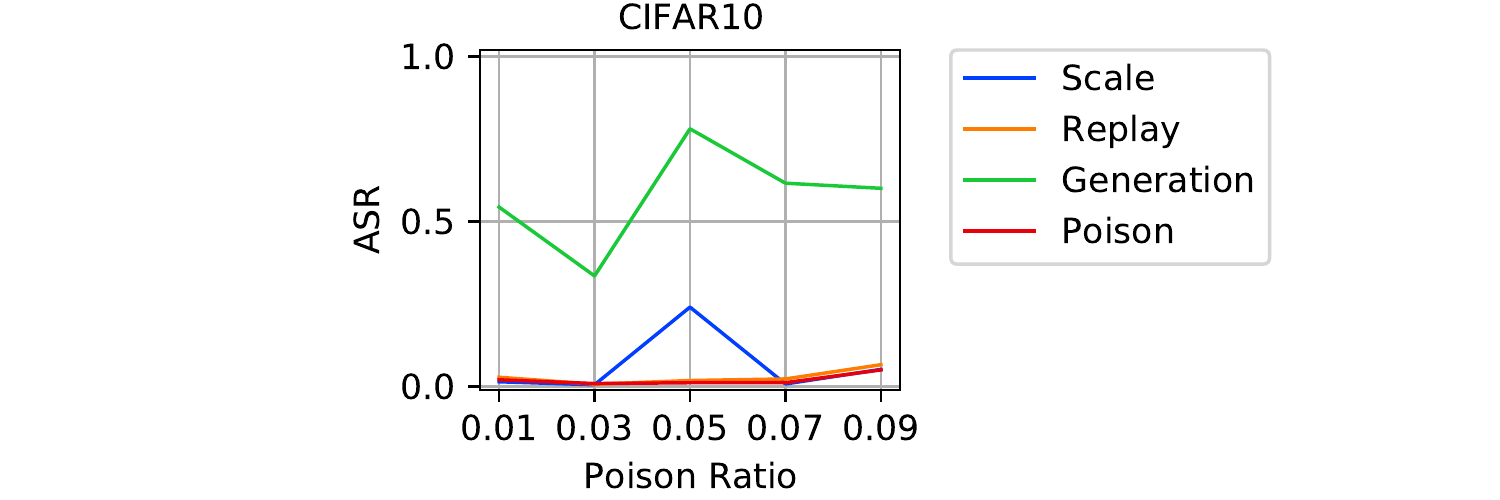}
    \caption{Effect of poison ratio. The x-axis denotes the proportion of poisoned samples among the target class's samples. The y-axis represents the ASR.}
    \label{fig:poison}
\end{figure}

This section evaluates the impact of the number of poisoned samples on our attacks using the CIFAR10 dataset, with the adversary's feature ratio fixed at 30\%. 
We vary the sampling ratio of poisoned samples from 1\% to 10\% of the target class's samples for comparison.
\pengyu{Fig.~\ref{fig:poison} plots the results of the experiments.}

As expected, increasing the number of samples added with $\textbf{r}$ enhances the performance of the poison attack, albeit with some limitations. For instance, at a poison ratio of 9\%, the poison attack achieves an ASR over 10\%, which is much better than the ASR of nearly 0\% at the poison ratio of 1\%.

Furthermore, the results show that as the $\textbf{r}$ becomes stronger, the performance of the generation attack, replay attack, and scale attack also improves, although with some fluctuations. 
Notably, the generation attack benefits more than the replay attack and the scale attack since it is initialized with $\textbf{r}$.

In conclusion, increasing the number of poisoned samples can enhance not only the performance of the poison attack but also the performance of other attacks. 
These findings also reveals the threat posed by malicious parties in the training of VFL models.

\subsection{Number of Queries}\label{sec:query}
\begin{figure}[t]
    \centering
    \setlength{\abovecaptionskip}{0.cm}
    \includegraphics[width=\columnwidth]{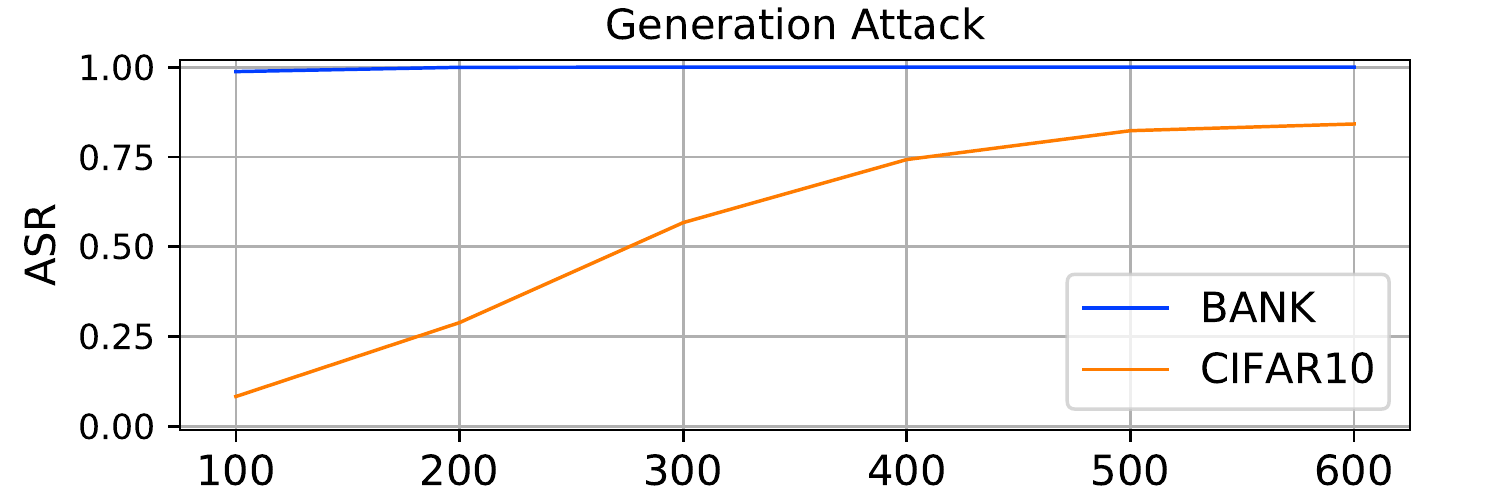}
    \caption{Performance with different query budgets. The x-axis represents the number of iterations. The y-axis represents the ASR.}
    \label{fig:query}
\end{figure}

This section examines the impact of the query budget on our attacks using the BANK and CIFAR10 datasets, with the adversary's feature ratio set at 30\%. 
We vary the number of iterations from 100 to 600, corresponding to the number of queries from 200 to 1,200.
\pengyu{Fig.~\ref{fig:query} plots the results of the experiments.}

The results show that for BANK, the generation attack achieves an ASR of almost 100\% with only 100 iterations, showing no significant trend with the increase in the number of queries. 
However, for CIFAR10, the attack's performance improves with an increasing number of queries until it converges when the number of iterations reaches 500. 
This phenomenon is due to the feature completeness and the number of classes as we discussed in Section~\ref{sec:ap}, which require more queries to generate a powerful $\textbf{r}$ for CIFAR10. 
Nevertheless, the improvement in performance has a limit, likely because the adversary lacks sufficient features.

Overall, our findings indicate that the generation attack's performance improves with an increasing number of queries, but there is a limit to the improvement due to the adversary's feature ratio.

\subsection{Impact of Modals}\label{sec:mm}
The characteristics of VFL make it well-suited for multi-modal learning. 
Thus, in this section, we evaluate the attacks in a multi-modal scenario with $\mathcal{K}=(\mathcal{P},\mathcal{L}_t)$ using the AP dataset, which includes both text and tabular data. 

The text data contains the description of an APP, while the tabular data records the characteristics of an APP, such as the number of permissions and the security level of requested permissions. 
Intuitively, the tabular data is much more critical for classification than the text data, implying that attacks targeting the tabular data should perform better.

The results in TABLE~\ref{tab:modal} support this intuition. 
When the adversary has access to the tabular data, the scale attack, replay attack, and generation attack achieve an ASR of 100\%. 
However, for the text data, only the generation attacks can achieve the same performance. 
For the other three attacks, the importance of the text data for classification is limited, resulting in a performance lower than 20\%. 
This finding reconfirms our speculation that a participant with more important data for the top model can carry out a more potent attack.

In conclusion, our attacks are agnostic to the modality of the data. 
The attack's capability is determined by the importance of the features to the task at hand.

\begin{table}[t]
\centering
\caption{Comparison of different data modal. The value in the cell denotes the ASR of the corresponding attack and the data modal.}
\label{tab:modal}
\resizebox{\columnwidth}{!}{%
\begin{tabular}{@{}lcccc@{}}
\toprule
Modal   & \multicolumn{1}{l}{Poison Attack} & \multicolumn{1}{l}{Scale Attack} & \multicolumn{1}{l}{Replay Attack} & \multicolumn{1}{l}{Generation Attack} \\ \midrule
Tabular & 0.93\%                            & 100\%                            & 100\%                             & 100\%                                 \\
Text    & 15.02\%                           & 11.23\%                          & 22.05\%                           & 100\%                                 \\ \bottomrule
\end{tabular}%
}
\end{table}

\subsection{Running Time}\label{sec:rt}
This section assesses the running time of our attacks with $\mathcal{K}=(\mathcal{P},\times)$. 
We fix the feature ratio at 10\%, and it does not affect the attack procedure.

\textbf{Attack Complexity:}
The scale attack and replay attack have a theoretical complexity of $O(N)$, where $N$ is the size of the training set. 
The scale attack requires samples from the target class and randomly selects them, leading to the need to search the training set for enough samples.
Similarly, the replay attack needs samples with high confidence from the target class, thus given a threshold, e.g., 0.9, the complexity does not exceed $O(N)$.

For the generation attack, the complexity of the selection part is also $O(N)$ since it involves finding samples from the source class. 
The subsequent procedure is to compute the differentiation of two queries, which takes $O(Q)$ time, where $Q$ is the query budget. 
However, since $Q$ is limited to a small constant in our cases, we conclude that the generation attack's complexity is also $O(N)$.

\textbf{Experimental Results:}
TABLE~\ref{tab:rt} reports the running time of different attacks on various datasets. 
Interestingly, the results show no significant difference in the time spent for different attacks on the same dataset. This outcome occurs because the timer records the period, including computing posteriors and calculating the ASR. Therefore, the model structure's complexity dominates the time cost rather than the attack itself. 
For example, Bert takes much longer to compute the activation values than ResNet and MLP.

The difference in running time between the scale attack and the replay attack is due to the implementation. 
We use the `random' package to generate a value and select the sample if the value is over 0.9 for the scale attack. Therefore, the scale attack takes slightly more time to select enough samples than the replay attack, especially when finding high confidence samples is easy.

In summary, our experimental results demonstrate that the running time of the attacks is not significantly affected by their theoretical complexity but rather the model structure's complexity, which accounts for most of the computation time.

\begin{table}[t]
\centering
\caption{Running time (s) of different kinds of attacks on each dataset.}
\label{tab:rt}
\resizebox{\columnwidth}{!}{%
\begin{tabular}{@{}lllllll@{}}
\toprule
Attacks           & BANK & CRITEO & IMDB     & MNIST & CIFAR10 & FER   \\ \midrule
Scale Attack      & 2.10 & 14.21  & 1047.19  & 30.55 & 40.58   & 79.45 \\
Replay Attack     & 1.01 & 11.96  & 1385.05  & 28.40 & 39.55   & 67.70  \\
Generation Attack & 4.29 & 16.52  & 1339.01  & 31.75 & 42.31   & 67.89 \\ \bottomrule
\end{tabular}%
}
\end{table}

\subsection{Memory Usage}\label{sec:mu}
This section evaluates the memory usage of different attacks with the same experimental setup as Section~\ref{sec:rt}. 
However, recording the memory usage of a function is challenging since an object's size may increase and decrease during the process.
To address this issue, we use the \textit{memory\_profile} and \textit{psutil} packages to measure the memory usage.

The memory usage results are summarized in TABLE~\ref{tab:mu}, which includes the training data, i.e., images and texts, loaded in the attack. 
Each attack consumes almost the same memory for each dataset, with slight differences possibly resulting from intermediate calculation results' storage.

However, note that different running environments, such as Windows and Linux, have different scheduling logic that can affect memory usage. 
Therefore, our results in this section should be taken as reference.

\begin{table}[t]
\centering
\caption{Memory usage (MiB) of different kinds of attacks on each dataset.}
\label{tab:mu}
\resizebox{\columnwidth}{!}{%
\begin{tabular}{@{}lllllll@{}}
\toprule
Attacks           & BANK   & CRITEO & IMDB   & MNIST  & CIFAR10 & FER    \\ \midrule
Scale Attack      & 2463.1 & 2510.1 & 2478.8 & 3120.7 & 3369.9  & 3041.9 \\
Replay Attack     & 2463.2 & 2510.3 & 2478.9 & 3120.9 & 3370.1  & 3042.1 \\
Generation Attack & 2463.4 & 2510.4 & 2479.0 & 3121.8 & 3370.3  & 3042.3 \\ \bottomrule
\end{tabular}%
}
\end{table}

\section{Defense}
In this section, our focus is on exploring potential defense strategies against the attacks we have identified. 
Since we are introducing a new track of attack against VFL, there is a lack of existing research dedicated to developing defense mechanisms for this specific context. 
The core challenges that arise when designing defense methods for VFL can be summarized as follows:
\begin{itemize}
    \item Due to privacy protection measures, at inference time, the adversary's embedding in plaintext is not available for detection, leaving only posterior information.
    \item The assumption of `clean' datasets is no longer valid since the adversary's cooperation is required in preparing the validation and test datasets, allowing the adversary to poison any `clean' datasets.
\end{itemize}
In addition, as our attacks rely on exploiting features with a strong signal, we are seeking defense methods that can be incorporated into the training process, such as a robust model design.

In the following section, we will begin by analyzing existing defenses for traditional centralized scenarios from two streams: adversarial perturbation generation and poisoning attacks. 
We will then provide an explanation for why we have chosen two specific methods as references for our defense strategies. 
Finally, the experimental results will be presented in Section~\ref{sec:norm} and Section~\ref{sec:drop}.

\subsection{Analysis of Existing Defenses}
Since our attacks involves feature generation and poisoning phase, we investigate defenses on these two perspectives for inspiration.

\textbf{Defenses against Adversarial Attacks:}
Firstly, defenses such as \cite{wang2019neural,grosse2017statistical,pang2018detection} that utilize clean datasets to count statistical properties of benign samples for detection are not applicable in this scenario.

After excluding the above-mentioned detection methods, existing defense strategies mainly focus on adversarial training \cite{shafahi2019adversarial,ganin2016domain} and input transformation \cite{guo2017countering,mustafa2019adversarial} to obtain a robust model. 
However, adversarial training still requires all participants to cooperate, making it vulnerable to attacks by adversaries. 
Additionally, the computational and communication resources required for adversarial training make it impractical for real business scenarios.

Therefore, we have chosen input transformation as one of our defense methods. 
According to findings in \cite{liu2018fine,bagdasaryan2020backdoor}, limiting the size of activations can be helpful for defending against numerical attacks. 
To achieve this goal, we normalize $\textbf{v}_i$ from each participant following the method proposed in \cite{mustafa2019adversarial}.

\textbf{Defenses against Poisoning Attacks:}
To begin, it should be noted that our attacks are orthogonal to poisoning attacks. 
We mainly combine the poisoning phase to perturb the top model's weights and strengthen our attacks. 
Furthermore, based on previous evaluations, we have found that the implanted feature alone cannot modify the model's prediction. 
Therefore, we speculate that existing defenses may not be suitable for our attacks.

Specifically, according to \cite{gao2020backdoor}, existing defenses can be classified into two categories: offline defense and online detection. 
Offline defense requires data and model inspection. 
For example, Neural Cleanse \cite{wang2019neural}, a typical method in this category, attempts to generate the trigger reversely. 
However, due to the performance of the $\textbf{r}$ and the incomplete access to the model, it is not feasible. 
On the other hand, online detection methods such as Artificial Brain Stimulation (ABS) \cite{Liu2019abs} and STRIP \cite{gao2019strip} require a set of clean samples' normal `behavior' to determine the threshold for identifying abnormalities, which is impractical due to the second challenge.

In \cite{Ilyas2019features}, Ilyas et al. indicated that useful but non-robust features could cause misclassification. 
We speculate that our planted features plays the same role, and the top model may memorize its connection with the target class. 
Therefore, we introduce dropout \cite{Srivastava2014dropout} to reduce the top model's memory.

In summary, we have adopted two countermeasures, normalization and dropout, to defend against our attacks. 
The following sections will evaluate these methods separately.

\subsection{Normalization}\label{sec:norm}
In this section, we will evaluate the impact of normalization on our attacks. Since we use ZOO to approximate gradients of a single coordinate, this strategy will significantly weaken our ability. Additionally, the performance of the scale attack will also be affected by normalization on each sample's embedding.

\begin{figure}[t]
    \centering
    \setlength{\abovecaptionskip}{0.cm}
    \includegraphics[width=\columnwidth]{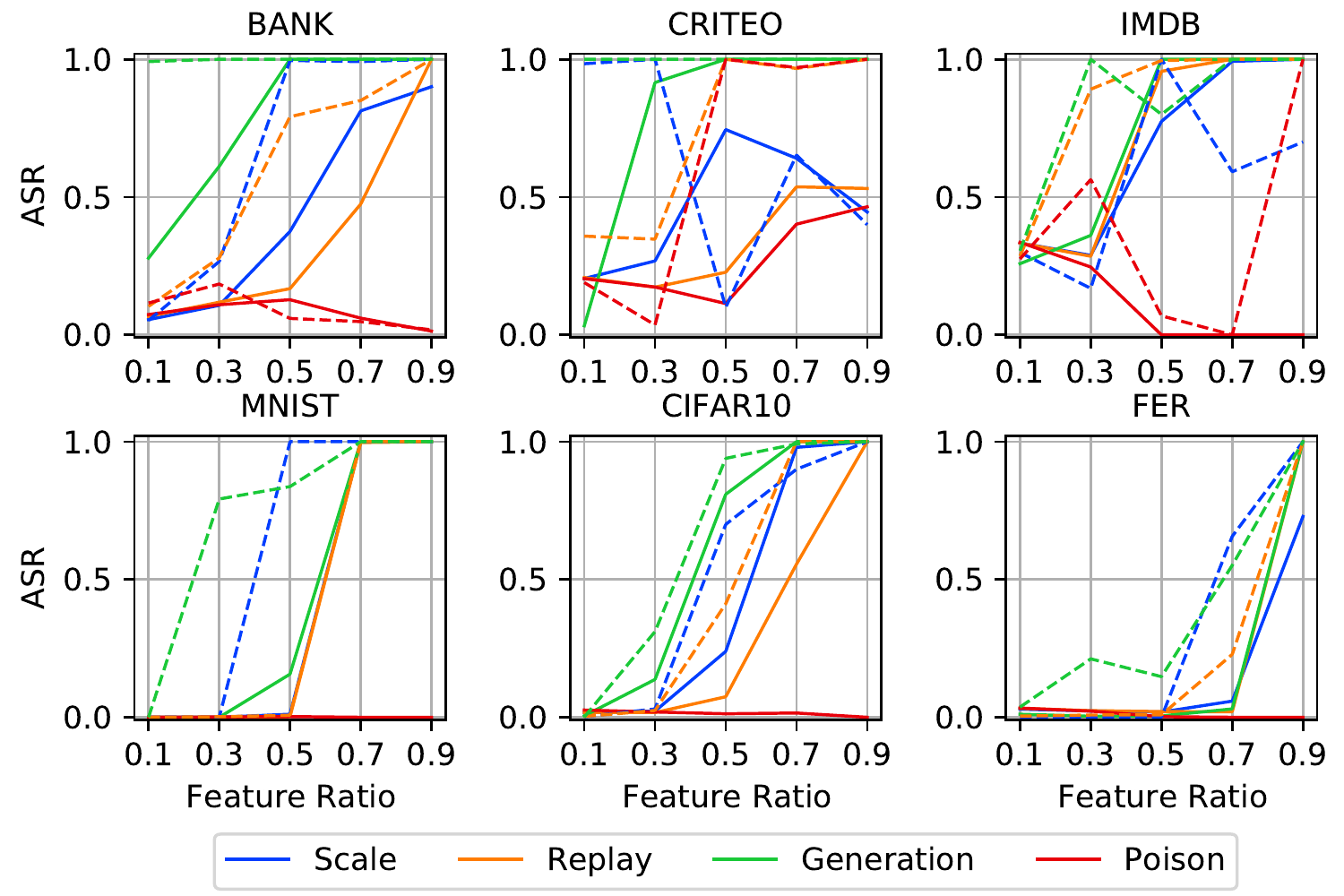}
    \caption{Impact of normalization on attack performance. The x-axis represents the feature ratio of the adversary. The y-axis denotes the ASR. The dashed line denotes the performance without defense.}
    \label{fig:norm}
\end{figure}

We conducted experiments using the same settings as in Section~\ref{sec:apt}, where $\mathcal{K}=(\mathcal{P},\mathcal{L})$, and varied the feature ratio of the adversary for comparison. Fig.~\ref{fig:norm} shows the results of our attacks' performance under normalization. 
The impact is evident on all datasets. For example, when the feature ratio is 10\%, our generation attack can only achieve an ASR of 30\% on BANK, which is over 99\% in Section~\ref{sec:apt}. 
Furthermore, our attacks fail on MNIST when the feature ratio is 30\%, and the generation attack drops from 85\% to 0\%.
A similar drop occurs in the scale attack and the replay attack.

Despite the drop in performance at small feature ratios, our attacks' performance remains when the feature ratio is over 50\%, indicating that the threat still exists.

In conclusion, normalization does affect the efficiency of our attacks when the adversary's feature ratio is small. However, its impact is limited when the adversary has sufficient features, making him/her important to the final predictions. 
The vulnerability still lies in the fact that the current VFL lacks design for fault tolerance.

\subsection{Dropout}\label{sec:drop}
This section evaluates the impact of dropout as a defense mechanism. 
We added one dropout layer to the top model after the activation layer (e.g., ReLU) and varied the probability of dropout from 0.1 to 0.9 for comparison. 
The probability denotes how likely a neuron's value will be set to zero. 
Other settings were the same as those in Section~\ref{sec:apt}, and we set the feature ratio to 30\%.

\begin{figure}[t]
    \centering
    \setlength{\abovecaptionskip}{0.cm}
    \includegraphics[width=\columnwidth]{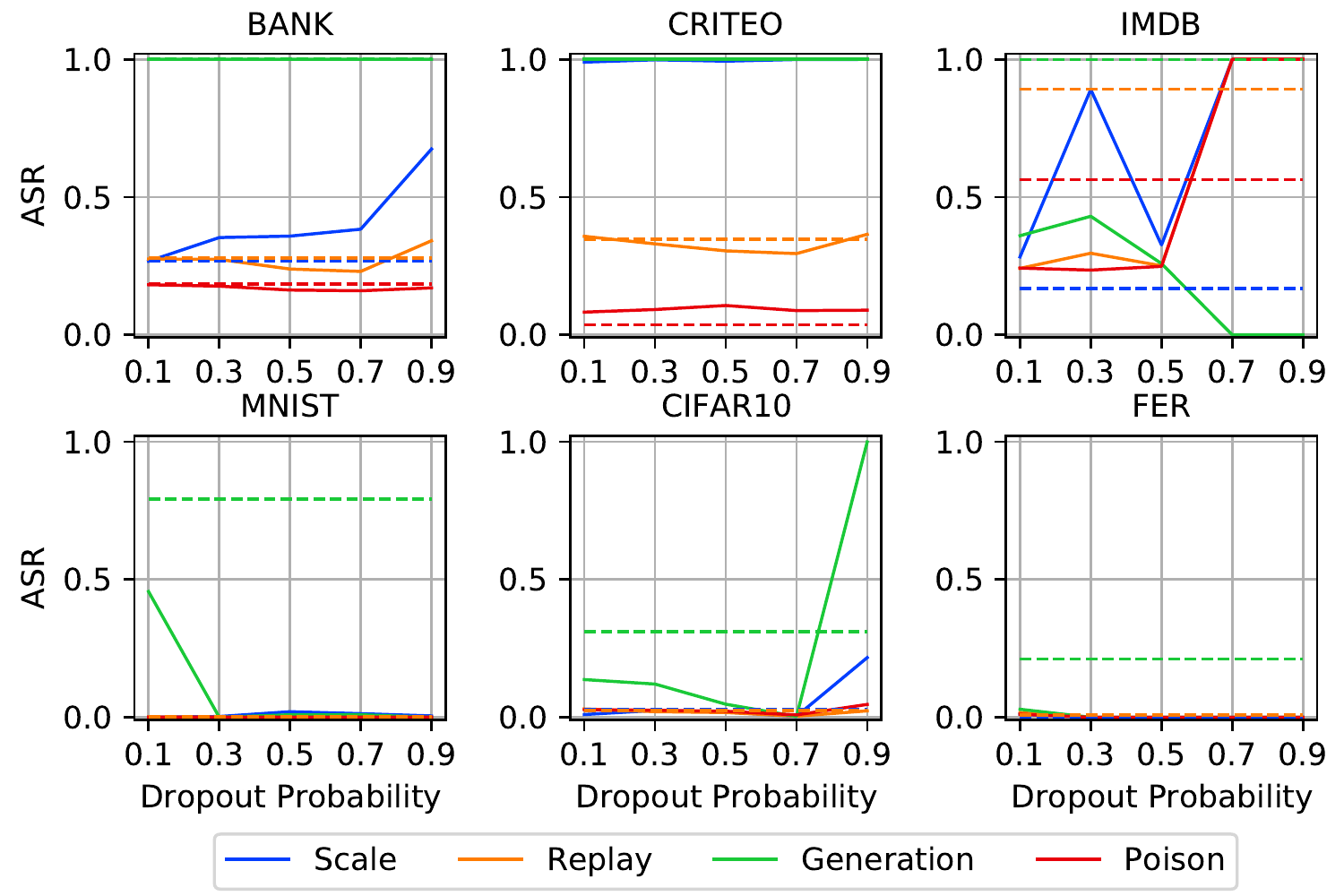}
    \caption{Impact of dropout on attack performance. The x-axis represents the feature ratio of the adversary. The y-axis denotes the ASR. The dashed line denotes the performance without defense.}
    \label{fig:dropout}
\end{figure}

Fig.~\ref{fig:dropout} shows the results. 
For BANK and CRITEO, the impact of dropout is minimal. 
Furthermore, the scale attack and the replay attack even perform better when the probability is set to 0.9. 
We speculate that this may be because dropout also severely weakens the availability of features of the other party, making the importance of the adversary relatively increase. 
A similar phenomenon occurs on IMDB, although there is more fluctuation.

For MNIST, CIFAR10, and FER, there is no change in the scale attack and the replay attack. 
However, the generation attack almost fails on these datasets. 
To understand the reason for this failure, we further examin the experiments' details. 
We find that the accuracy of the main task rapidly decreases as the probability of dropout increases. 
Thus, we speculate that the low accuracy implies that the top model loses its ability to predict precisely.
As a result, our $\textbf{v}_{adv}$ also loses its effectiveness since the top model behaves like a random guess.

In conclusion, dropout can reduce the model's memorization of specific patterns, which is helpful. 
However, with a high probability of dropout, not only does the attack's efficiency decrease, but the main task's performance drops rapidly. 
In VFL, this performance loss is unacceptable, as it violates the motivation of collaborative learning.

\section{Related Work}
\subsection{Attacks and Defenses in FL}
Federated Learning (FL) is rapidly gaining popularity in the industry, leading to an increase in the development of attacks and defenses aimed at ensuring its secure application. 
Currently, much of the research in FL is focused on the scenario of Horizontal Federated Learning (HFL), where multiple clients share the same feature space but have different sample spaces. 
The security and privacy implications of VFL have yet to be widely studied. 
In the following, we primarily provide an overview of the attacks and defenses of HFL, while also presenting the limited available research on VFL to our best.

\textbf{Attack Vectors:}
In HFL, participants have the same feature space $\mathcal{F}$ but different sample spaces $\mathcal{U}$. 
The aim of HFL is to get a global model by aggregating the parameters of bottom models, which are trained from each participant's data.
Therefore, in this scenario, studying adversarial attacks is almost the same as centralized learning.
As a result, much of the research in this area has focused on backdoor attacks against HFL.

The goal of the backdoor attack in HFL is to implant a trigger in the global model, allowing an adversary to manipulate the results of the model users. 
Bagdasaryan et al. were the first to propose such an attack in \cite{bagdasaryan2020backdoor}. 
They implemented a simple but effective method for injecting a backdoor by scaling up the gradients. 
Subsequent research by Sun et al. \cite{sun2019can} and Wang et al. \cite{wang2020attack} explored the efficacy of backdoor attacks in more complex scenarios, such as the EMNIST dataset. 
Xie et al. \cite{xie2019dba} proposed a distributed implementation of the attack by splitting the triggers into multiple parts.

Liu et al. \cite{liu2020backdoor} were the first to study the backdoor attack against VFL. 
They used the gradients of one sample belonging to the target class to replace the target samples, allowing them to bypass the limitation of not being able to change labels during training. 
However, their experimental results showed that their method had a negative impact on the performance of the original task and was not effective on the selected datasets.

Indeed, these works highlight the importance of designing FL, both HFL and VFL, to be resilient to malicious parties. 
It is crucial to consider potential security and privacy threats and to develop effective countermeasures to ensure the integrity and reliability of FL systems.

\textbf{Defense Strategies:}
To counter the attack vectors mentioned earlier, several defense strategies for HFL have been proposed. 
These strategies focus on detecting malicious updates from clients and feedback from the global model.

In \cite{fung2018mitigating}, Fung et al. proposed a defense mechanism that identifies poisoning attacks based on the diversity of client updates during the distributed learning process. 
This system is an improvement over prior work as it does not impose constraints on the expected number of adversaries and requires fewer assumptions about clients and their data. 
In \cite{li2020learning}, Li et al. proposed a new framework for robust federated learning, where the central server uses a powerful detection model to identify and remove malicious model updates.

BAFFLE, proposed in \cite{andreina2021baffle}, is a method for detecting backdoor attacks through feedback-based federated learning. 
The authors leveraged the data from multiple clients to not only train the model but also uncover model poisoning. 
Additionally, Wu et al. \cite{wu2021mitigating} proposed a federated pruning method that removes redundant neurons in the network and adjusts the extreme weight values of the model to mitigate attacks.

In their work \cite{BELHADI2021fault,djenouri2023fault}, the authors proposed a method for detecting secure faults in the Industrial Internet of Things using HFL. 
They identified anomalous patterns in time series data at the local level, and then used a reinforcement learning-based merging strategy to securely aggregate these uploads. 
Singh et al. also proposed a novel scheme called FusionFedBlock in \cite{SINGH2023fedblock}, which addresses centralization, privacy preservation, latency, and security in the Industrial Internet of Things, leaving secure aggregation as the job of miners in the blockchain.

Gao et al. considered a scenario in which the server is malicious in their work \cite{GAO2023sverifl}. 
For instance, the malicious server may conduct dishonest data aggregation and return incorrect aggregated gradients to all participants. 
To address this problem, they proposed SVeriFL, which enables the integrity of parameters uploaded by participants and the correctness of aggregated results of the server to be verified.

Although the defenses in HFL impressively mitigate the threat of existing attack vectors, they are not suitable for VFL, especially for our attacks. 
This is because, in HFL, the purpose of the defenses is to obtain a clean model, whereas in VFL, our attacks aim to induce the top model to give desired predictions at inference time, where the models have already been employed.

\subsection{Adversarial Attacks against DNNs}
The adversarial attack happens at inference phase, where an adversary crafts a specific perturbation to fool the target DNN.
According to the coverage of the perturbation, the adversarial attack can be divided into two categories: sample-related adversarial attack and universal adversarial attack.

\textbf{Sample-related Adversarial Attack:} The sample-related adversarial attack refers to the kind of attack whose generated perturbation is correlated to the base sample.
For example, in image classification, given an input image, $\textbf{x}$, e.g., `cat', and the target DNN model $f$, the adversary can change $f(\textbf{x})$'s prediction from `cat' to whatever he/she desires, e.g., `dog', with an adversarial sample $\textbf{x}_\delta$, which is obtained from the formula $\mathcal{B}_\epsilon(\textbf{x})=\left\{\textbf{x}_\delta~|~||\textbf{x}_\delta-\textbf{x}||_p\leq \epsilon\right\}$, where $\mathcal{B}_\epsilon(\textbf{x})$ denotes a norm ball at $\textbf{x}$, $p$ denotes the $L_p$ norm, and $\epsilon$ denotes the bound.
To make the attack stealthy, the adversary often restricts $\epsilon$ to a small value.
Formally, the generation of such an adversarial attack is formulated as the optimization objective:
\begin{equation}\label{eq:adversarial}
    \mathop{\min}_{\textbf{x}_\delta\in\mathcal{B}_\epsilon(\textbf{x})} \ell(\textbf{x}_\delta,t;\theta),
\end{equation}
where $\ell$ measures the loss of the adversarial sample's prediction with the target class $y_t$; $\theta$ is the parameter of $f$.

Many works have studied the optimization of Eq.~(\ref{eq:adversarial}). 
According to the adversary's knowledge about $\theta$, there are two primary categories: white-box and black-box attacks.  
In the white-box scenario, the adversary knows the model's structure and parameters.
Therefore, the perturbation can be optimized by the gradient descent method, such as FGSM \cite{goodfellow2014explaining}, PGD \cite{madry2017towards}, and C\&W \cite{carlini2017towards}. 
In the black-box scenario, the adversary only has access to the model's output, which means he/she cannot obtain the gradients.
To address this, there are two streams for approximating the gradients. 
One class is the differential method \cite{Chen2017zoo}, which calculates the perturbation's gradients through the difference of two queries. 
The other class tries to approximate the gradients by a surrogate model \cite{li2020learning}, which is trained on an auxiliary dataset that is independently and identically distributed (i.i.d) to the training dataset. 

\textbf{Universal Adversarial Attacks:}
The universal adversarial attack refers to the kind of attack whose generated perturbation can work on a group of samples, namely Universal Adversarial Perturbation (UAP).

In \cite{moosavidezfooli2017universal}, Moosavidez-Dezfooli et al. found the existence of a universal (image-agnostic) and small perturbation vector that causes natural images to be misclassified with high probability.
However, the UAP generated by their method only aims to cause misclassification of the model without a target class.
Therefore, it is not applicable for a malicious adversary who has the same goal in our work.

In \cite{brown2018adversarial}, Brown et al. proposed a more powerful attack to create universal, robust, targeted adversarial image patches. 
Their evaluation results show that with careful design of the loss function, the generated adversarial patches can have an impressive attack performance.
Similarly, in \cite{goh2021multimodal}, Goh et al. found the neurons that are sensitive to specific features.
Specifically, they showed that putting a card that printed `iPod' on a Granny Smith apple would make their model misclassify it as iPod rather than an apple.

\subsection{Poisoning Attacks against DNNs}
This kind of attack aims to pollute DNNs during the training phase. 
Previous works \cite{jagielski2018manipulating,alfeld2016data} focus on decreasing the model's utility, e.g., making the model unable to converge in training by modifying the training data.
Recently, a line of works \cite{jagielski2020subpopulation,ali2018poison,bagdasaryan2020backdoor} finds that the model can also be controlled by a fixed pattern without sacrifice the main task's performance, called backdoor attack.
According to the control of labeling process, the backdoor attacks can also be classified into two streams.

\textbf{Label-controlled Backdoor Attack:} This kind of attacks aims to perturb the model's parameters $\theta$, denoted by $\theta_\delta$, with modifying a set of samples' labels.
To ensure the evasiveness of the attack, the adversary should further limit the perturbation on $\theta$ in a given function space $\mathcal{F}$, which is defined by $\mathcal{F}_\epsilon(\theta)=\left\{\theta_\delta~|~\mathbb{E}_{\textbf{x}\in D_{val}}[|f(\textbf{x};\theta_\delta)-f(\textbf{x};\theta)|\leq\epsilon]\right\}$, where $D_{val}$ refers to the validation dataset, and $\epsilon$ denotes the error bound of the performance loss.
Then, the objective function becomes as:
\begin{equation}\label{eq:poison}
    \mathop{\min}_{\theta_\delta\in\mathcal{F}_\epsilon(\theta)}\mathbb{E}_{\textbf{x}\in D_{poisoned}}[\ell(\textbf{x},y_t;\theta_\delta)],
\end{equation}
where $D_{poisoned}$ denotes the set of poisoned inputs, $y_t$ is the target label, and the loss function is defined similarly to Eq.~(\ref{eq:adversarial}).

Optimizing Eq.~(\ref{eq:poison}) in practice can be challenging. 
As a result, many studies have focused on polluting training data \cite{jagielski2020subpopulation,liu2018Trojannn} to induce deviation in DNN parameters or modifying DNN structures \cite{tang2020embarrassingly}. 
In the former type of attack, the adversary limits perturbations to $\theta$ by reducing the number of poisoned samples. 
In the latter type of attack, the adversary modifies neurons or adds extra malicious structures to DNN models.

\textbf{Clean-label Backdoor Attack:}
In \cite{ali2018poison}, the researchers believed that an adversary could not control the labeling process, and even if he/she does, the mislabelled samples are easy to recognize.
Therefore, they proposed a more stealthy method via feature collision.
That is, the adversary first picks up a sample from the desired class, e.g., an image of `cat', and then adds an invisible perturbation to it.
The perturbation is used to make this poisoned sample's latent embedding as close to that of the target instance (e.g., an image of `dog') as possible.
Then, when the target instance is being inferred, it will be misclassified as `cat'.

Formally, let $f$ be the model without the last softmax layer, $\textbf{x}$ be the sample from the desired class, and $\textbf{x}_t$ be the target sample. A poisoning sample $\textbf{x}_p$ is crafted as follows:
\begin{equation}\label{eq:clean}
    \mathop{\min}_{\textbf{x}_p} \left\|f(\textbf{x}_p)-f(\textbf{x}_t)\right\|_2^2 + \beta \left\|\textbf{x}_p-\textbf{x}\right\|_2^2,
\end{equation}
where $\beta$ is a hyperparameter that controls the similarity between the poisoning sample $\textbf{x}_p$ and original sample $\textbf{x}$.
Usually, the clean-label attack needs a number of poisoning samples for one target instance.

\subsection{Remark}
\begin{table*}[t]
\small
\centering
\caption{Summary and comparison of existing work and our work. In the `Attack Vector' column, two existing attack vectors and ours are presented. In the `Attack Scene' column, `CL' denotes Centralized Learning, `HFL' denotes Horizontal Federated Learning, and `VFL' denotes Vertical Federated Learning. In the `Attack Assumption' column, the attacks are further categorize according to their assumptions. In the `Attack Phase' column, the attacking phase is presented. In the `Attack Coverage' column, we summarize the target coverage of the attack. Adversarial attacks typically target a specific sample, while other attacks aim to a group of samples. In the `Attack Operation' column, it provides a brief introduction to the corresponding attack operation. Finally, in the `References' column, representative relevant work are presented.}
\label{tab:summary}
\resizebox{\textwidth}{!}{%
\begin{tabular}{@{}ccccccc@{}}
\toprule
\textbf{Attack Vector} &
  \textbf{Attack Scene} &
  \textbf{Attack Assumption} &
  \textbf{Attack Phase} &
  \textbf{Attack Coverage} &
  \textbf{Attack Operation} &
  \multicolumn{1}{l}{\textbf{References}} \\ \midrule
\multirow{2}{*}{Adversarial Attack} &
  \multirow{2}{*}{CL} &
  White-box: Target model's parameters &
  Inference &
  Single sample &
  Add adversarial perturbation &
  \cite{goodfellow2014explaining, dong2018boosting, moosavi2016deepfool, moosavi2017universal} \\
     &     & Black-box: Target model's output                          & Inference & Single sample    & Add adversarial perturbation & \cite{Chen2017zoo, ilyas18blackbox, brown2018adversarial, Su2019pixel} \\ \midrule
\multirow{3}{*}{Backdoor Attack} &
  \multirow{2}{*}{CL} &
  Label-controlled: Polluting dataset through modifying samples and labels &
  Training &
  Group of samples &
  Add trigger &
  \cite{jagielski2020subpopulation,liu2018Trojannn,tang2020embarrassingly} \\
     &     & Clean-label: Polluting dataset through modifying samples  & Training  & Group of samples & Add trigger                  & \cite{ali2018poison,turner2019cleanlabel} \\
     & HFL & Control the client model's parameters used in aggregation & Training  & Group of samples & Replace uploaded parameters  & \cite{bagdasaryan2020backdoor,xie2019dba} \\ \midrule
Ours & VFL & Control the bottom model's output used in aggregation     & Inference & Group of samples & Replace uploaded features    & \\ \bottomrule
\end{tabular}%
}
\end{table*}

Our work is the first to explore how to manipulate VFL model's predictions by replacing features as one party.
To enhance the effectiveness of our attacks, we draw upon valuable insights from the domains of adversarial and poisoning attacks. 
Specifically, we adopt an optimization-based approach to generate features, inspired by adversarial attacks. 
We also enhance the power of the features by increasing their frequency in samples of the target class, inspired by poisoning attacks.
However, directly transferring these attacks to VFL does not work. 
The following are the limitations of these attacks. 

\textbf{Limitation of Adversarial Attacks:} Existing adversarial attacks \cite{goodfellow2014explaining, dong2018boosting, moosavi2016deepfool, moosavi2017universal, Chen2017zoo, ilyas18blackbox, brown2018adversarial, Su2019pixel, Li2019text} typically assume a centralized setting where the adversary has access to complete features. 
White-box attacks \cite{goodfellow2014explaining, dong2018boosting, moosavi2016deepfool, moosavi2017universal, Li2019text} further assume full access to the model, which is not applicable in VFL. 
In contrast, black-box attacks \cite{Chen2017zoo, ilyas18blackbox, brown2018adversarial, Su2019pixel} are designed to address the challenge of limited access to the model, relying on posteriors to guide optimization. 
For image classification, \cite{brown2018adversarial, Su2019pixel} have demonstrated success in minimizing the number of modified pixels, even down to a single pixel.

However, in VFL, the adversary is constrained to modifying only their own features, which may not be given sufficient importance in the model's prediction. 
Additionally, our attacks aim to operate on a class of samples with a single perturbation, which falls outside the scope of the above attacks.

\textbf{Limitation of Backdoor Attacks:} In scenarios where the adversary owns the labels (typically the initiator of a VFL model), poisoning attacks are easy to implement, and their success rates are consistent with those in the centralized setting \cite{liu2020backdoor}. 
Hence, we do not discuss this scenario as it is relatively trivial.

The adversary can also be the party who provides only features \cite{webankvflcase1,webankvflcase2}. 
For example, in Fig.~\ref{fig:vfl}, a financial company is paid to provide user information for the bank. 
In this case, backdoor attacks that rely on modifying the labels of poisoning samples are not applicable, as no label information is available.
Although the clean-label backdoor attack does not modify the labels, the adversary's partial control of features makes it much harder to manipulate a class of samples' embeddings to the target instance's position. 

TABLE~\ref{tab:summary} shows the difference of existing work and our work, which is also a brief summary of above discussion.
In summary, our work highlights that in VFL, the challenge of the BGP is much more severe than expected, as the adversary can leverage both adversarial and poisoning attacks to enhance their abilities.

\section{Discussion}
\subsection{Bypass Normalization}
The evaluation of layer normalization's impact on our attack shows it is an efficient way to defend black-box attacks like ZOO.
The key to its success is to transform the perturbed input so that the adversary cannot approximate the gradients.
Two possible attack strategies may solve the challenge. 
One is based on the perturbation's transferability across models \cite{papernot2016transferability}. 
Another uses the heuristic search for generating perturbations \cite{brendel2018decisionbased}. 

Brendel et al. \cite{brendel2018decisionbased} suggested that adversarial sample generation can be achieved through heuristic search, and similar ideas are explored in other works such as \cite{wang2020mgaattack,taori2019targeted}. 
We plan to investigate these approaches in future work.

\subsection{Integrated Defense}
The defenses we evaluated were proposed for traditional scenarios and targeted specific types of attacks. 
However, our experiments demonstrate that even if a defense successfully prevents one aspect of our attack, such as perturbation generation, it may still be vulnerable to another aspect, such as poisoning attacks. 
This observation suggests a new direction for defense design, where we should consider both attack vectors to provide more robust protection for DNN-based systems. 
By taking a holistic approach to defense, we may be able to create defenses that can withstand a wider range of attacks and provide better overall security.

\subsection{Consensus in VFL}
In VFL, the lack of a clear definition of consensus distinguishes it from well-defined BGPs. 
Typically, the main task's performance during training serves as an indicator of the participants' honesty and determines whether the model's prediction is reliable. 
However, this leaves the security issue to be addressed at inference time. 
Specifically, when a consensus is made, such as the top model's predictions, it becomes challenging to determine whether the model's prediction is correct without human intervention. 
Moreover, in a two-party scenario, it is difficult to distinguish whether a participant is cheating if the prediction is incorrect.

To address this issue, a promising approach may be to pre-define clustering centers for each class during training, as suggested by Hoe et al. \cite{hoe2021one}. 
At inference time, embeddings whose distance from the center is larger than a threshold could be judged as invalid. 
We plan to investigate the design of a new VFL framework that incorporates this approach in our future work.
\section{Conclusion}\label{sec:conclusion}
Our work exposes the dangers of malicious manipulation of VFL models by a single party using two attacks, i.e., the replay attack and the generation attack. 
Additionally, the poisoning phase demonstrates how the adversary's power can increase by perturbing the top model's parameters.
We also investigate existing defense methods and assess two strategies that can be incorporated during training to counteract our attacks. 
Experimental results reveal the effectiveness of our attacks and emphasize the urgency to develop advanced defense techniques to safeguard VFL models.



\section*{Acknowledgments}
This work was partly supported by the National Key Research and Development Program of China under No. 2022YFB3102100 and NSFC under No. 62102360.

\ifCLASSOPTIONcaptionsoff
  \newpage
\fi


\bibliographystyle{IEEEtranS}
\bibliography{references}

\begin{thebibliography}{10}
\providecommand{\url}[1]{#1}
\csname url@samestyle\endcsname
\providecommand{\newblock}{\relax}
\providecommand{\bibinfo}[2]{#2}
\providecommand{\BIBentrySTDinterwordspacing}{\spaceskip=0pt\relax}
\providecommand{\BIBentryALTinterwordstretchfactor}{4}
\providecommand{\BIBentryALTinterwordspacing}{\spaceskip=\fontdimen2\font plus
\BIBentryALTinterwordstretchfactor\fontdimen3\font minus
  \fontdimen4\font\relax}
\providecommand{\BIBforeignlanguage}[2]{{%
\expandafter\ifx\csname l@#1\endcsname\relax
\typeout{** WARNING: IEEEtranS.bst: No hyphenation pattern has been}%
\typeout{** loaded for the language `#1'. Using the pattern for}%
\typeout{** the default language instead.}%
\else
\language=\csname l@#1\endcsname
\fi
#2}}
\providecommand{\BIBdecl}{\relax}
\BIBdecl

\bibitem{abdel2014convolutional}
O.~Abdel-Hamid, A.-r. Mohamed, H.~Jiang, L.~Deng, G.~Penn, and D.~Yu,
  ``Convolutional neural networks for speech recognition,'' \emph{IEEE/ACM
  Transactions on audio, speech, and language processing}, vol.~22, no.~10, pp.
  1533--1545, 2014.

\bibitem{afriat2020capitalism}
H.~Afriat, S.~Dvir-Gvirsman, K.~Tsuriel, and L.~Ivan, ``“this is capitalism.
  it is not illegal”: Users’ attitudes toward institutional privacy
  following the cambridge analytica scandal,'' \emph{The Information Society},
  vol.~37, no.~2, pp. 115--127, 2020.

\bibitem{alfeld2016data}
S.~Alfeld, X.~Zhu, and P.~Barford, ``Data poisoning attacks against
  autoregressive models,'' in \emph{Proceedings of the AAAI Conference on
  Artificial Intelligence}, vol.~30, no.~1, 2016.

\bibitem{andreina2021baffle}
S.~Andreina, G.~A. Marson, H.~M{\"o}llering, and G.~Karame, ``Baffle: Backdoor
  detection via feedback-based federated learning,'' in \emph{2021 IEEE 41st
  International Conference on Distributed Computing Systems (ICDCS)}.\hskip 1em
  plus 0.5em minus 0.4em\relax IEEE, 2021, pp. 852--863.

\bibitem{bagdasaryan2020backdoor}
E.~Bagdasaryan, A.~Veit, Y.~Hua, D.~Estrin, and V.~Shmatikov, ``How to backdoor
  federated learning,'' in \emph{International Conference on Artificial
  Intelligence and Statistics}.\hskip 1em plus 0.5em minus 0.4em\relax PMLR,
  2020, pp. 2938--2948.

\bibitem{Barsoum2016fer}
E.~Barsoum, C.~Zhang, C.~Canton~Ferrer, and Z.~Zhang, ``Training deep networks
  for facial expression recognition with crowd-sourced label distribution,'' in
  \emph{ACM International Conference on Multimodal Interaction (ICMI)}, 2016.

\bibitem{BELHADI2021fault}
A.~Belhadi, Y.~Djenouri, G.~Srivastava, A.~Jolfaei, and J.~C.-W. Lin, ``Privacy
  reinforcement learning for faults detection in the smart grid,'' \emph{Ad Hoc
  Networks}, vol. 119, p. 102541, 2021.

\bibitem{bertsekas1997nonlinear}
D.~P. Bertsekas, ``Nonlinear programming,'' \emph{Journal of the Operational
  Research Society}, vol.~48, no.~3, pp. 334--334, 1997.

\bibitem{brendel2018decisionbased}
W.~Brendel, J.~Rauber, and M.~Bethge, ``Decision-based adversarial attacks:
  Reliable attacks against black-box machine learning models,'' 2018.

\bibitem{brown2018adversarial}
T.~B. Brown, D.~Mané, A.~Roy, M.~Abadi, and J.~Gilmer, ``Adversarial patch,''
  2018.

\bibitem{carlini2017towards}
N.~Carlini and D.~Wagner, ``Towards evaluating the robustness of neural
  networks,'' in \emph{2017 ieee symposium on security and privacy (sp)}.\hskip
  1em plus 0.5em minus 0.4em\relax IEEE, 2017, pp. 39--57.

\bibitem{Chen2017zoo}
P.-Y. Chen, H.~Zhang, Y.~Sharma, J.~Yi, and C.-J. Hsieh, ``Zoo,''
  \emph{Proceedings of the 10th ACM Workshop on Artificial Intelligence and
  Security}, Nov 2017.

\bibitem{morten2018tfencrypted}
\BIBentryALTinterwordspacing
M.~Dahl, J.~Mancuso, Y.~Dupis, B.~Decoste, M.~Giraud, I.~Livingstone,
  J.~Patriquin, and G.~Uhma, ``Private machine learning in tensorflow using
  secure computation,'' \emph{CoRR}, vol. abs/1810.08130, 2018. [Online].
  Available: \url{http://arxiv.org/abs/1810.08130}
\BIBentrySTDinterwordspacing

\bibitem{devlin2019bert}
J.~Devlin, M.-W. Chang, K.~Lee, and K.~Toutanova, ``Bert: Pre-training of deep
  bidirectional transformers for language understanding,'' 2019.

\bibitem{djenouri2023fault}
Y.~Djenouri, A.~Belhadi, G.~Srivastava, U.~Ghosh, P.~Chatterjee, and J.~C.-W.
  Lin, ``Fast and accurate deep learning framework for secure fault diagnosis
  in the industrial internet of things,'' \emph{IEEE Internet of Things
  Journal}, vol.~10, no.~4, pp. 2802--2810, 2023.

\bibitem{dong2018boosting}
Y.~Dong, F.~Liao, T.~Pang, H.~Su, J.~Zhu, X.~Hu, and J.~Li, ``Boosting
  adversarial attacks with momentum,'' in \emph{Proceedings of the IEEE
  conference on computer vision and pattern recognition}, 2018, pp. 9185--9193.

\bibitem{fiesler2020no}
C.~Fiesler, N.~Beard, and B.~C. Keegan, ``No robots, spiders, or scrapers:
  Legal and ethical regulation of data collection methods in social media terms
  of service,'' in \emph{Proceedings of the international AAAI conference on
  web and social media}, vol.~14, 2020, pp. 187--196.

\bibitem{fu2022label}
C.~Fu, X.~Zhang, S.~Ji, J.~Chen, J.~Wu, S.~Guo, J.~Zhou, A.~X. Liu, and
  T.~Wang, ``Label inference attacks against vertical federated learning,'' in
  \emph{31st USENIX Security Symposium (USENIX Security 22)}.\hskip 1em plus
  0.5em minus 0.4em\relax Boston, MA: USENIX Association, Aug. 2022, pp.
  1397--1414.

\bibitem{fung2018mitigating}
C.~Fung, C.~J. Yoon, and I.~Beschastnikh, ``Mitigating sybils in federated
  learning poisoning,'' \emph{arXiv preprint arXiv:1808.04866}, 2018.

\bibitem{ganin2016domain}
Y.~Ganin, E.~Ustinova, H.~Ajakan, P.~Germain, H.~Larochelle, F.~Laviolette,
  M.~Marchand, and V.~Lempitsky, ``Domain-adversarial training of neural
  networks,'' \emph{The journal of machine learning research}, vol.~17, no.~1,
  pp. 2096--2030, 2016.

\bibitem{GAO2023sverifl}
H.~Gao, N.~He, and T.~Gao, ``Sverifl: Successive verifiable federated learning
  with privacy-preserving,'' \emph{Information Sciences}, vol. 622, pp.
  98--114, 2023.

\bibitem{gao2020backdoor}
Y.~Gao, B.~G. Doan, Z.~Zhang, S.~Ma, J.~Zhang, A.~Fu, S.~Nepal, and H.~Kim,
  ``Backdoor attacks and countermeasures on deep learning: A comprehensive
  review,'' \emph{arXiv preprint arXiv:2007.10760}, 2020.

\bibitem{gao2019strip}
Y.~Gao, C.~Xu, D.~Wang, S.~Chen, D.~C. Ranasinghe, and S.~Nepal, ``Strip: A
  defence against trojan attacks on deep neural networks,'' in
  \emph{Proceedings of the 35th Annual Computer Security Applications
  Conference}, ser. ACSAC '19.\hskip 1em plus 0.5em minus 0.4em\relax New York,
  NY, USA: Association for Computing Machinery, 2019, p. 113–125.

\bibitem{goh2021multimodal}
G.~Goh, N.~C. †, C.~V. †, S.~Carter, M.~Petrov, L.~Schubert, A.~Radford,
  and C.~Olah, ``Multimodal neurons in artificial neural networks,''
  \emph{Distill}, 2021, https://distill.pub/2021/multimodal-neurons.

\bibitem{goodfellow2014explaining}
I.~J. Goodfellow, J.~Shlens, and C.~Szegedy, ``Explaining and harnessing
  adversarial examples,'' \emph{arXiv preprint arXiv:1412.6572}, 2014.

\bibitem{grosse2017statistical}
K.~Grosse, P.~Manoharan, N.~Papernot, M.~Backes, and P.~McDaniel, ``On the
  (statistical) detection of adversarial examples,'' \emph{arXiv preprint
  arXiv:1702.06280}, 2017.

\bibitem{guo2017countering}
C.~Guo, M.~Rana, M.~Cisse, and L.~Van Der~Maaten, ``Countering adversarial
  images using input transformations,'' \emph{arXiv preprint arXiv:1711.00117},
  2017.

\bibitem{Guo2017criteo}
H.~Guo, R.~Tang, Y.~Ye, Z.~Li, and X.~He, ``Deepfm: A factorization-machine
  based neural network for ctr prediction,'' in \emph{IJCAI}, 2017.

\bibitem{he2015deep}
K.~He, X.~Zhang, S.~Ren, and J.~Sun, ``Deep residual learning for image
  recognition,'' 2015.

\bibitem{herdiani2021analysis}
F.~D. Herdiani, ``Analysis of abuse and fraud in the legal and illegal online
  loan fintech application using the hybrid method,'' \emph{Enrichment: Journal
  of Management}, vol.~11, no.~2, pp. 486--490, 2021.

\bibitem{hoe2021one}
J.~T. Hoe, K.~W. Ng, T.~Zhang, C.~S. Chan, Y.-Z. Song, and T.~Xiang, ``One loss
  for all: Deep hashing with a single cosine similarity based learning
  objective,'' in \emph{Advances in Neural Information Processing Systems},
  M.~Ranzato, A.~Beygelzimer, Y.~Dauphin, P.~Liang, and J.~W. Vaughan, Eds.,
  vol.~34.\hskip 1em plus 0.5em minus 0.4em\relax Curran Associates, Inc.,
  2021, pp. 24\,286--24\,298.

\bibitem{ilyas18blackbox}
A.~Ilyas, L.~Engstrom, A.~Athalye, and J.~Lin, ``Black-box adversarial attacks
  with limited queries and information,'' in \emph{Proceedings of the 35th
  International Conference on Machine Learning}, ser. Proceedings of Machine
  Learning Research, J.~Dy and A.~Krause, Eds., vol.~80.\hskip 1em plus 0.5em
  minus 0.4em\relax PMLR, 10--15 Jul 2018, pp. 2137--2146.

\bibitem{Ilyas2019features}
A.~Ilyas, S.~Santurkar, D.~Tsipras, L.~Engstrom, B.~Tran, and A.~Madry,
  ``Adversarial examples are not bugs, they are features,'' in \emph{Advances
  in Neural Information Processing Systems}, H.~Wallach, H.~Larochelle,
  A.~Beygelzimer, F.~d\textquotesingle Alch\'{e}-Buc, E.~Fox, and R.~Garnett,
  Eds., vol.~32.\hskip 1em plus 0.5em minus 0.4em\relax Curran Associates,
  Inc., 2019.

\bibitem{iwaya2018mhealth}
L.~H. Iwaya, S.~Fischer-H{\"u}bner, R.-M. {\AA}hlfeldt, and L.~A. Martucci,
  ``mhealth: A privacy threat analysis for public health surveillance
  systems,'' in \emph{2018 IEEE 31st International Symposium on Computer-Based
  Medical Systems (CBMS)}.\hskip 1em plus 0.5em minus 0.4em\relax IEEE, 2018,
  pp. 42--47.

\bibitem{jagielski2018manipulating}
M.~Jagielski, A.~Oprea, B.~Biggio, C.~Liu, C.~Nita-Rotaru, and B.~Li,
  ``Manipulating machine learning: Poisoning attacks and countermeasures for
  regression learning,'' in \emph{2018 IEEE Symposium on Security and Privacy
  (SP)}.\hskip 1em plus 0.5em minus 0.4em\relax IEEE, 2018, pp. 19--35.

\bibitem{jagielski2020subpopulation}
M.~Jagielski, G.~Severi, N.~P. Harger, and A.~Oprea, ``Subpopulation data
  poisoning attacks,'' \emph{arXiv preprint arXiv:2006.14026}, 2020.

\bibitem{kiselbach2012new}
D.~Kiselbach and C.~E. Joern, ``New consumer product safety laws in canada and
  the united states: Business on the border,'' \emph{Global Trade and Customs
  Journal}, vol.~7, no.~1, 2012.

\bibitem{ref:crypten}
B.~Knott, S.~Venkataraman, A.~Hannun, S.~Sengupta, M.~Ibrahim, and L.~van~der
  Maaten, ``Crypten: Secure multi-party computation meets machine learning,''
  in \emph{arXiv 2109.00984}, 2021.

\bibitem{krizhevsky2009learning}
A.~Krizhevsky, G.~Hinton \emph{et~al.}, ``Learning multiple layers of features
  from tiny images,'' Toronto, ON, Canada, Tech. Rep., 2009.

\bibitem{krizhevsky2012imagenet}
A.~Krizhevsky, I.~Sutskever, and G.~E. Hinton, ``Imagenet classification with
  deep convolutional neural networks,'' \emph{Advances in neural information
  processing systems}, vol.~25, pp. 1097--1105, 2012.

\bibitem{lamport1985solve}
\BIBentryALTinterwordspacing
L.~Lamport, ``Solved problems, unsolved problems and non-problems in
  concurrency,'' \emph{SIGOPS Oper. Syst. Rev.}, vol.~19, no.~4, p. 34–44,
  oct 1985. [Online]. Available: \url{https://doi.org/10.1145/858336.858339}
\BIBentrySTDinterwordspacing

\bibitem{lamport2019byzantine}
\BIBentryALTinterwordspacing
L.~Lamport, R.~Shostak, and M.~Pease, \emph{The Byzantine Generals
  Problem}.\hskip 1em plus 0.5em minus 0.4em\relax New York, NY, USA:
  Association for Computing Machinery, 2019, p. 203–226. [Online]. Available:
  \url{https://doi.org/10.1145/3335772.3335936}
\BIBentrySTDinterwordspacing

\bibitem{lax2014calculus}
P.~D. Lax and M.~S. Terrell, \emph{Calculus with applications}.\hskip 1em plus
  0.5em minus 0.4em\relax Springer, 2014.

\bibitem{lecun1998gradient}
Y.~LeCun, L.~Bottou, Y.~Bengio, and P.~Haffner, ``Gradient-based learning
  applied to document recognition,'' \emph{Proceedings of the IEEE}, vol.~86,
  no.~11, pp. 2278--2324, 1998.

\bibitem{Li2019text}
J.~Li, S.~Ji, T.~Du, B.~Li, and T.~Wang, ``{TextBugger}: Generating adversarial
  text against real-world applications,'' in \emph{Proceedings 2019 Network and
  Distributed System Security Symposium}.\hskip 1em plus 0.5em minus
  0.4em\relax Internet Society, 2019.

\bibitem{li2020learning}
S.~Li, Y.~Cheng, W.~Wang, Y.~Liu, and T.~Chen, ``Learning to detect malicious
  clients for robust federated learning,'' \emph{arXiv preprint
  arXiv:2002.00211}, 2020.

\bibitem{LIANG2016561}
D.~Liang, C.-C. Lu, C.-F. Tsai, and G.-A. Shih, ``Financial ratios and
  corporate governance indicators in bankruptcy prediction: A comprehensive
  study,'' \emph{European Journal of Operational Research}, vol. 252, no.~2,
  pp. 561--572, 2016.

\bibitem{liu2018fine}
K.~Liu, B.~Dolan-Gavitt, and S.~Garg, ``Fine-pruning: Defending against
  backdooring attacks on deep neural networks,'' in \emph{Research in Attacks,
  Intrusions, and Defenses: 21st International Symposium, RAID 2018, Heraklion,
  Crete, Greece, September 10-12, 2018, Proceedings 21}.\hskip 1em plus 0.5em
  minus 0.4em\relax Springer, 2018, pp. 273--294.

\bibitem{liu2021fate}
Y.~Liu, T.~Fan, T.~Chen, Q.~Xu, and Q.~Yang, ``Fate: An industrial grade
  platform for collaborative learning with data protection,'' \emph{Journal of
  Machine Learning Research}, vol.~22, no. 226, pp. 1--6, 2021.

\bibitem{liu2019communication}
Y.~Liu, Y.~Kang, X.~Zhang, L.~Li, Y.~Cheng, T.~Chen, M.~Hong, and Q.~Yang, ``A
  communication efficient vertical federated learning framework,'' \emph{arXiv
  preprint arXiv:1912.11187}, 2019.

\bibitem{liu2020backdoor}
Y.~Liu, Z.~Yi, and T.~Chen, ``Backdoor attacks and defenses in
  feature-partitioned collaborative learning,'' 2020.

\bibitem{Liu2021BatchLI}
\BIBentryALTinterwordspacing
Y.~Liu, T.~Zou, Y.~Kang, W.~Liu, Y.~He, Z.~qian Yi, and Q.~Yang, ``Batch label
  inference and replacement attacks in black-boxed vertical federated
  learning,'' 2021. [Online]. Available: \url{https://arxiv.org/abs/2112.05409}
\BIBentrySTDinterwordspacing

\bibitem{Liu2019abs}
Y.~Liu, W.-C. Lee, G.~Tao, S.~Ma, Y.~Aafer, and X.~Zhang, ``Abs: Scanning
  neural networks for back-doors by artificial brain stimulation,'' in
  \emph{Proceedings of the 2019 ACM SIGSAC Conference on Computer and
  Communications Security}, ser. CCS '19.\hskip 1em plus 0.5em minus
  0.4em\relax New York, NY, USA: Association for Computing Machinery, 2019, p.
  1265–1282.

\bibitem{liu2018Trojannn}
Y.~Liu, S.~Ma, Y.~Aafer, W.-C. Lee, J.~Zhai, W.~Wang, and X.~Zhang, ``Trojaning
  attack on neural networks,'' in \emph{25th Annual Network and Distributed
  System Security Symposium, {NDSS} 2018, San Diego, California, USA, February
  18-221, 2018}.\hskip 1em plus 0.5em minus 0.4em\relax The Internet Society,
  2018.

\bibitem{Luo2021feature}
X.~Luo, Y.~Wu, X.~Xiao, and B.~C. Ooi, ``Feature inference attack on model
  predictions in vertical federated learning,'' in \emph{37th {IEEE}
  International Conference on Data Engineering, {ICDE} 2021, Chania, Greece,
  April 19-22, 2021}.\hskip 1em plus 0.5em minus 0.4em\relax {IEEE}, 2021, pp.
  181--192.

\bibitem{maas2011word}
A.~L. Maas, R.~E. Daly, P.~T. Pham, D.~Huang, A.~Y. Ng, and C.~Potts,
  ``Learning word vectors for sentiment analysis,'' in \emph{Proceedings of the
  49th Annual Meeting of the Association for Computational Linguistics: Human
  Language Technologies}.\hskip 1em plus 0.5em minus 0.4em\relax Portland,
  Oregon, USA: Association for Computational Linguistics, June 2011, pp.
  142--150.

\bibitem{madry2017towards}
A.~Madry, A.~Makelov, L.~Schmidt, D.~Tsipras, and A.~Vladu, ``Towards deep
  learning models resistant to adversarial attacks,'' \emph{arXiv preprint
  arXiv:1706.06083}, 2017.

\bibitem{arvind2018android}
A.~Mahindru, ``Android permission dataset,'' 2018.

\bibitem{moosavi2017universal}
S.-M. Moosavi-Dezfooli, A.~Fawzi, O.~Fawzi, and P.~Frossard, ``Universal
  adversarial perturbations,'' 2017.

\bibitem{moosavidezfooli2017universal}
------, ``Universal adversarial perturbations,'' 2017.

\bibitem{moosavi2016deepfool}
S.-M. Moosavi-Dezfooli, A.~Fawzi, and P.~Frossard, ``Deepfool: a simple and
  accurate method to fool deep neural networks,'' in \emph{Proceedings of the
  IEEE conference on computer vision and pattern recognition}, 2016, pp.
  2574--2582.

\bibitem{mustafa2019adversarial}
A.~Mustafa, S.~Khan, M.~Hayat, R.~Goecke, J.~Shen, and L.~Shao, ``Adversarial
  defense by restricting the hidden space of deep neural networks,'' in
  \emph{Proceedings of the IEEE/CVF International Conference on Computer
  Vision}, 2019, pp. 3385--3394.

\bibitem{pang2018detection}
T.~Pang, C.~Du, Y.~Dong, and J.~Zhu, ``Towards robust detection of adversarial
  examples,'' in \emph{Advances in Neural Information Processing Systems},
  S.~Bengio, H.~Wallach, H.~Larochelle, K.~Grauman, N.~Cesa-Bianchi, and
  R.~Garnett, Eds., vol.~31.\hskip 1em plus 0.5em minus 0.4em\relax Curran
  Associates, Inc., 2018.

\bibitem{papernot2016transferability}
N.~Papernot, P.~McDaniel, and I.~Goodfellow, ``Transferability in machine
  learning: from phenomena to black-box attacks using adversarial samples,''
  2016.

\bibitem{qiu2022relation}
P.~Qiu, X.~Zhang, S.~Ji, T.~Du, Y.~Pu, J.~Zhou, and T.~Wang, ``Your labels are
  selling you out: Relation leaks in vertical federated learning,'' \emph{IEEE
  Transactions on Dependable and Secure Computing}, pp. 1--16, 2022.

\bibitem{ali2018poison}
A.~Shafahi, W.~R. Huang, M.~Najibi, O.~Suciu, C.~Studer, T.~Dumitras, and
  T.~Goldstein, ``Poison frogs! targeted clean-label poisoning attacks on
  neural networks,'' in \emph{Proceedings of the 32nd International Conference
  on Neural Information Processing Systems}, ser. NIPS'18.\hskip 1em plus 0.5em
  minus 0.4em\relax Red Hook, NY, USA: Curran Associates Inc., 2018, p.
  6106–6116.

\bibitem{shafahi2019adversarial}
A.~Shafahi, M.~Najibi, A.~Ghiasi, Z.~Xu, J.~Dickerson, C.~Studer, L.~S. Davis,
  G.~Taylor, and T.~Goldstein, ``Adversarial training for free!'' \emph{arXiv
  preprint arXiv:1904.12843}, 2019.

\bibitem{SINGH2023fedblock}
S.~K. Singh, L.~T. Yang, and J.~H. Park, ``Fusionfedblock: Fusion of blockchain
  and federated learning to preserve privacy in industry 5.0,''
  \emph{Information Fusion}, vol.~90, pp. 233--240, 2023.

\bibitem{Srivastava2014dropout}
N.~Srivastava, G.~Hinton, A.~Krizhevsky, I.~Sutskever, and R.~Salakhutdinov,
  ``Dropout: A simple way to prevent neural networks from overfitting,''
  \emph{J. Mach. Learn. Res.}, vol.~15, no.~1, p. 1929–1958, Jan. 2014.

\bibitem{Su2019pixel}
J.~Su, D.~V. Vargas, and K.~Sakurai, ``One pixel attack for fooling deep neural
  networks,'' \emph{{IEEE} Transactions on Evolutionary Computation}, vol.~23,
  no.~5, pp. 828--841, oct 2019.

\bibitem{sun2019can}
Z.~Sun, P.~Kairouz, A.~T. Suresh, and H.~B. McMahan, ``Can you really backdoor
  federated learning?'' \emph{arXiv preprint arXiv:1911.07963}, 2019.

\bibitem{szegedy2013deep}
C.~Szegedy, A.~Toshev, and D.~Erhan, ``Deep neural networks for object
  detection,'' in \emph{Advances in Neural Information Processing Systems},
  C.~Burges, L.~Bottou, M.~Welling, Z.~Ghahramani, and K.~Weinberger, Eds.,
  vol.~26.\hskip 1em plus 0.5em minus 0.4em\relax Curran Associates, Inc.,
  2013.

\bibitem{Tang2016ExtremeLM}
J.~Tang, C.~Deng, and G.~Huang, ``Extreme learning machine for multilayer
  perceptron,'' \emph{IEEE Transactions on Neural Networks and Learning
  Systems}, vol.~27, pp. 809--821, 2016.

\bibitem{tang2020embarrassingly}
R.~Tang, M.~Du, N.~Liu, F.~Yang, and X.~Hu, ``An embarrassingly simple approach
  for trojan attack in deep neural networks,'' in \emph{Proceedings of the 26th
  ACM SIGKDD International Conference on Knowledge Discovery \& Data Mining},
  2020, pp. 218--228.

\bibitem{taori2019targeted}
R.~Taori, A.~Kamsetty, B.~Chu, and N.~Vemuri, ``Targeted adversarial examples
  for black box audio systems,'' in \emph{2019 IEEE Security and Privacy
  Workshops (SPW)}.\hskip 1em plus 0.5em minus 0.4em\relax IEEE, 2019, pp.
  15--20.

\bibitem{tenney2019bert}
I.~Tenney, D.~Das, and E.~Pavlick, ``Bert rediscovers the classical nlp
  pipeline,'' \emph{arXiv preprint arXiv:1905.05950}, 2019.

\bibitem{turner2019cleanlabel}
A.~Turner, D.~Tsipras, and A.~Madry, ``Clean-label backdoor attacks,'' 2019.

\bibitem{valiant1984pac}
L.~G. Valiant, ``A theory of the learnable,'' \emph{Commun. ACM}, vol.~27,
  no.~11, p. 1134–1142, nov 1984.

\bibitem{voigt2017eu}
P.~Voigt and A.~Von~dem Bussche, ``The eu general data protection regulation
  (gdpr),'' \emph{A Practical Guide, 1st Ed., Cham: Springer International
  Publishing}, 2017.

\bibitem{wang2019neural}
B.~Wang, Y.~Yao, S.~Shan, H.~Li, B.~Viswanath, H.~Zheng, and B.~Y. Zhao,
  ``Neural cleanse: Identifying and mitigating backdoor attacks in neural
  networks,'' in \emph{2019 IEEE Symposium on Security and Privacy (SP)}.\hskip
  1em plus 0.5em minus 0.4em\relax IEEE, 2019, pp. 707--723.

\bibitem{guan2019shapley}
G.~Wang, ``Interpret federated learning with shapley values,'' 2019.

\bibitem{wang2020attack}
H.~Wang, K.~Sreenivasan, S.~Rajput, H.~Vishwakarma, S.~Agarwal, J.-y. Sohn,
  K.~Lee, and D.~Papailiopoulos, ``Attack of the tails: Yes, you really can
  backdoor federated learning,'' \emph{arXiv preprint arXiv:2007.05084}, 2020.

\bibitem{wang2020mgaattack}
L.~Wang, K.~Yang, W.~Wang, R.~Wang, and A.~Ye, ``Mgaattack: Toward more
  query-efficient black-box attack by microbial genetic algorithm,'' in
  \emph{Proceedings of the 28th ACM International Conference on Multimedia},
  2020, pp. 2229--2236.

\bibitem{webankvflcase1}
Webank, ``A case of traffic violations insurance-using federated learning,''
  2020, \url{https://www.fedai.org/cases}.

\bibitem{webankvflcase2}
------, ``Utilization of {FATE} in risk management of credit in small and micro
  enterprises,'' 2020, \url{https://www.fedai.org/cases}.

\bibitem{weng2021privacy}
H.~Weng, J.~Zhang, F.~Xue, T.~Wei, S.~Ji, and Z.~Zong, ``Privacy leakage of
  real-world vertical federated learning,'' 2021.

\bibitem{wu2021mitigating}
C.~Wu, X.~Yang, S.~Zhu, and P.~Mitra, ``Mitigating backdoor attacks in
  federated learning,'' 2021.

\bibitem{xie2019dba}
C.~Xie, K.~Huang, P.-Y. Chen, and B.~Li, ``Dba: Distributed backdoor attacks
  against federated learning,'' in \emph{International Conference on Learning
  Representations}, 2019.

\bibitem{yang2019federated}
Q.~Yang, Y.~Liu, T.~Chen, and Y.~Tong, ``Federated machine learning: Concept
  and applications,'' \emph{ACM Transactions on Intelligent Systems and
  Technology (TIST)}, vol.~10, no.~2, pp. 1--19, 2019.

\bibitem{zhang2018mixup}
H.~Zhang, M.~Cisse, Y.~N. Dauphin, and D.~Lopez-Paz, ``mixup: Beyond empirical
  risk minimization,'' 2018.

\bibitem{zhou2021vertically}
J.~Zhou, C.~Chen, L.~Zheng, H.~Wu, J.~Wu, X.~Zheng, B.~Wu, Z.~Liu, and L.~Wang,
  ``Vertically federated graph neural network for privacy-preserving node
  classification,'' 2021.

\bibitem{ziller2021pysyft}
A.~Ziller, A.~Trask, A.~Lopardo, B.~Szymkow, B.~Wagner, E.~Bluemke, J.-M.
  Nounahon, J.~Passerat-Palmbach, K.~Prakash, N.~Rose, T.~Ryffel, Z.~N. Reza,
  and G.~Kaissis, \emph{PySyft: A Library for Easy Federated Learning}.\hskip
  1em plus 0.5em minus 0.4em\relax Cham: Springer International Publishing,
  2021, pp. 111--139.

\end{thebibliography}

\begin{IEEEbiography}[{\includegraphics[width=1in,height=1.25in,clip,keepaspectratio]{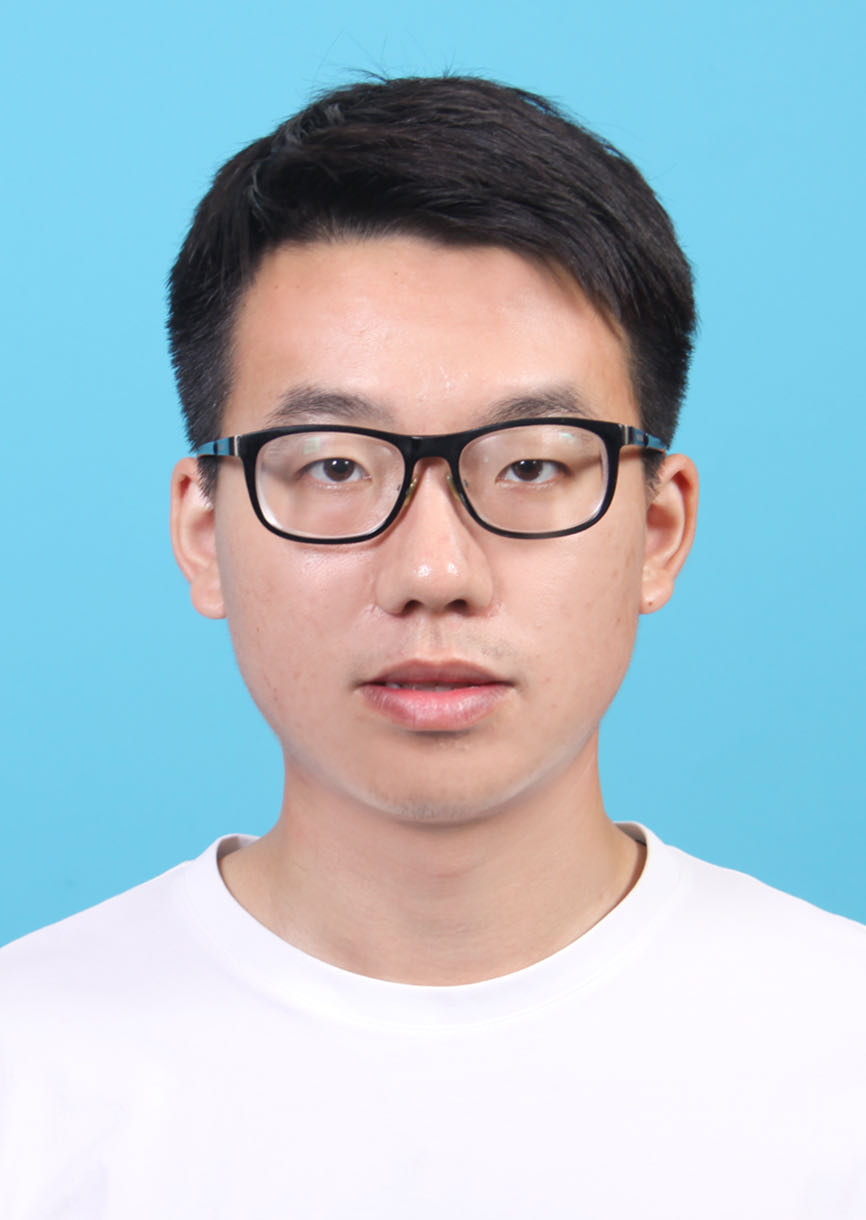}}]{Pengyu Qiu}
is currently a Ph.D. student in the College of Computer Science and Technology at Zhejiang University. He received his Bachelor's degree from Zhejiang University. His current research interests include AI security, adversarial learning.
\end{IEEEbiography}

\begin{IEEEbiography}[{\includegraphics[width=1in,height=1.25in,clip,keepaspectratio]{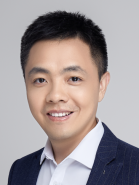}}]{Xuhong Zhang}
is a ZJU 100-Young Professor with the School of Software Technology at Zhejiang University. He received his Ph.D. in Computer Engineering from University of Central Florida in 2017 . His research interests include distributed big data and AI systems, big data mining and analysis, data-driven security, AI and Security. He has authored over 20 publications in premier journals and conferences such as TDSC, TPDC, IEEE S\&P, USENIX Security, ACM CCS, NDSS, VLDB, etc.
\end{IEEEbiography}

\begin{IEEEbiography}[{\includegraphics[width=1in,height=1.25in,clip,keepaspectratio]{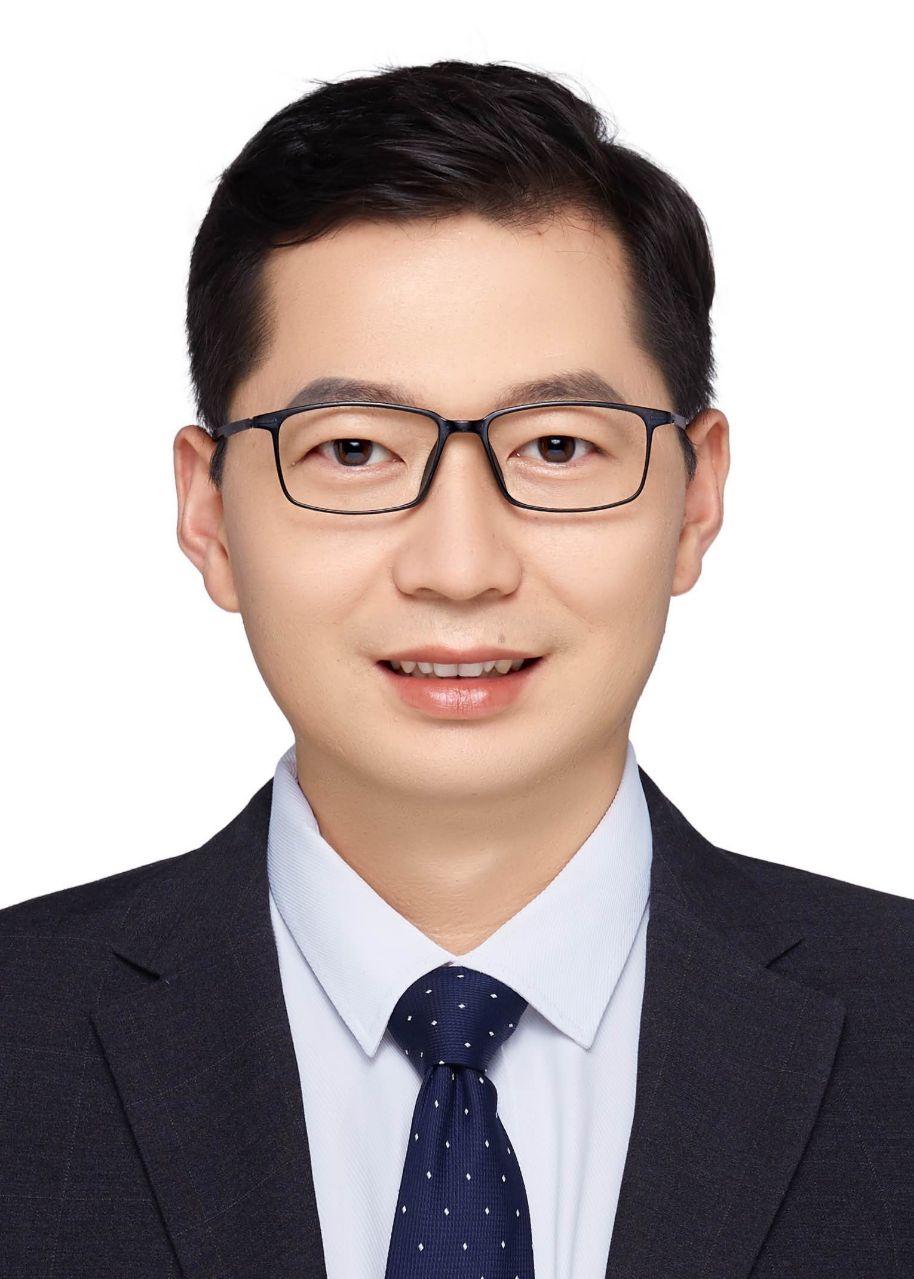}}]{Shouling Ji}
is a Qiushi Distinguished Professor in the College of Computer Science and Technology at Zhejiang University. He received a Ph.D. degree in Electrical and Computer Engineering from Georgia Institute of Technology and a Ph.D. degree in Computer Science from Georgia State University. His current research interests include Data-driven Security and Privacy, AI Security and Software and System Security. He is a member of ACM and IEEE, and a senior member of CCF. He was a Research Intern at the IBM T. J. Watson Research Center. Shouling is the recipient of the 2012 Chinese Government Award for Outstanding Self-Financed Students Abroad and 10 Best/Outstanding Paper Awards, including ACM CCS 2021.
\end{IEEEbiography}

\begin{IEEEbiography}[{\includegraphics[width=1in,height=1.25in,clip,keepaspectratio]{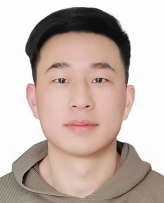}}]{Changjiang Li} 
is currently a Ph.D. student in the College of Information Science and Technology at Pennsylvania State University. He received the Master degree from the School of Computer Science at Zhejiang University in 2020. His research interest includes Adversarial Machine Learning, AI privacy.
\end{IEEEbiography}

\begin{IEEEbiography}[{\includegraphics[width=1in,height=1.25in,clip,keepaspectratio]{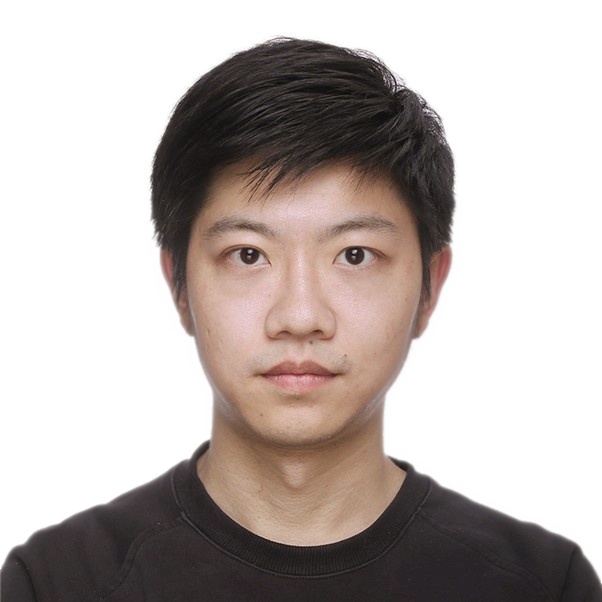}}]{Yuwen Pu} 
received his Ph.D. in the School of Big Data \& Software Engineering from Chongqing University in 2021. He is a PostDoctor in the College of Computer Science and Technology at Zhejiang University. His research interests include big data security and privacy-preserving, AI security.
\end{IEEEbiography}

\begin{IEEEbiography}[{\includegraphics[width=1in,height=1.25in,clip,keepaspectratio]{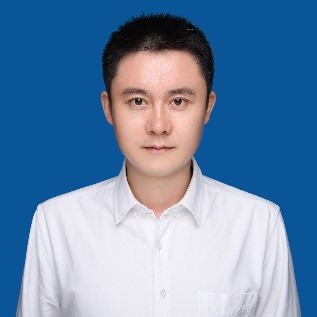}}]{Xing Yang} 
is a researcher at the State Key Laboratory of Pulsed Power Laser Technology, National University of Defense Technology. He received his BS, MS and Ph.D. degrees from Hefei Electronic Engineering Institute in 2006, 2009, and 2012 respectively. Currently, his research interests mainly focus on optoelectronic engineering, artificial intelligence, and cyberspace security.
\end{IEEEbiography}

\begin{IEEEbiography}[{\includegraphics[width=1in,height=1.25in,clip,keepaspectratio]{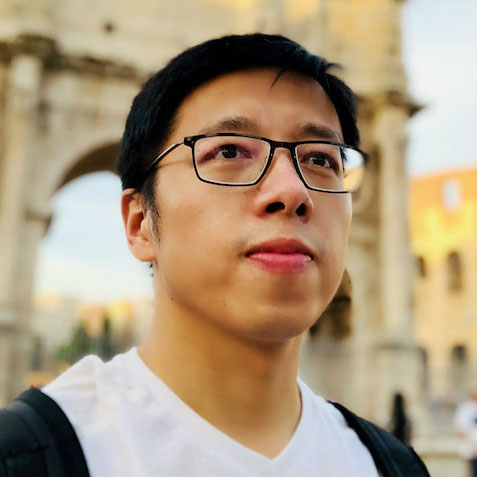}}]{Ting Wang}
is an assistant professor in the College of Information Sciences and Technology at Penn State. He received his Ph.D. degree from Georgia Tech. He conducts research at the intersection of data science and privacy \& security. His ongoing work focuses on making machine learning systems more practically usable through improving their Security Assurance, Privacy Preservation and Decision-Making Transparency.
\end{IEEEbiography}

\end{document}